\documentclass[10pt,journal,compsoc]{IEEEtran}

\ifCLASSOPTIONcompsoc
  \usepackage[nocompress]{cite}
\else
  \usepackage{cite}
\fi
\usepackage[pdftex]{graphicx}
\graphicspath{{./figs/}}
\DeclareGraphicsExtensions{.pdf,.jpg,.png}
\usepackage{cite}
\usepackage[cmex10]{amsmath}
\usepackage{amssymb}
\usepackage{graphicx}
\usepackage{cases}
\usepackage{mdwmath}
\usepackage{mdwtab}
\usepackage{array}
\usepackage{url}
\usepackage[ruled,vlined]{algorithm2e}
\usepackage{booktabs}
\usepackage{threeparttable}
\usepackage{color}
\usepackage{caption}
\usepackage{pifont}
\usepackage{wasysym}
\usepackage{amssymb}
\usepackage{multirow}
\usepackage{makecell}
\usepackage[colorlinks,linkcolor=black]{hyperref}    
\usepackage{marvosym}
\usepackage{lipsum}
\usepackage[caption=false,font=footnotesize]{subfig}
\usepackage{amsmath}
\usepackage{xspace}
\usepackage{bm}
\usepackage{amsthm,amsmath,amssymb}
\usepackage{mathrsfs}
\usepackage{mathrsfs}
\usepackage{multicol} 
\usepackage{hyperref}
\usepackage{tabularx}
\usepackage{longtable}
\usepackage{lipsum}

\hypersetup{
    colorlinks=true,             
    linkcolor=black,             
    filecolor=black,             
    urlcolor=black,              
    citecolor=black,             
    pdftitle={Overleaf Example},
    pdfpagemode=FullScreen,
}

\usepackage{hyperref}
\usepackage{overpic}

\usepackage{zhlipsum}
\usepackage{xcolor}

\usepackage[absolute,overlay]{textpos}

\makeatletter

\newcommand{\Rmnum}[1]{\expandafter\@slowromancap\romannumeral #1@}
\makeatother
\usepackage{graphicx}
\usepackage{float}
\usepackage{diagbox}

\newcommand{\ie}{\textit{i}.\textit{e}., }
\newcommand{\eg}{\textit{e}.\textit{g}., }
\def\etal{{\em et al.}}
\newcommand{\etc}{\textit{e}\textit{t}\textit{c}}

\ifCLASSINFOpdf
\else
\fi
\begin{document}
%
\title{A Comprehensive Survey on Segment Anything Model for Vision and Beyond}

\author{Chunhui~Zhang,
        Li~Liu$^*$,\IEEEmembership{~Member,~IEEE},
        Yawen~Cui,
        Guanjie~Huang,
        Weilin~Lin,
        Yiqian~Yang,
        Yuehong~Hu
\IEEEcompsocitemizethanks{
\IEEEcompsocthanksitem Chunhui~Zhang is with the Hong Kong University of Science and Technology (Guangzhou), Guangzhou 511458, China and Cooperative Medianet Innovation Center, Shanghai Jiao Tong University, Shanghai 200240, China and also with the CloudWalk Technology Co., Ltd, 201203, China. Email: chunhui.zhang@sjtu.edu.cn.\protect

\IEEEcompsocthanksitem Li~Liu is with the Hong Kong University of Science and Technology (Guangzhou), Guangzhou 511458, China. E-mail: liliu.math@gmail.com.\protect

\IEEEcompsocthanksitem Yawen~Cui is with the Hong Kong University of Science and Technology (Guangzhou), Guangzhou 511458, China and also with the University of Oulu. E-mail: yawen.cui@oulu.fi.\protect


\IEEEcompsocthanksitem Guanjie~Huang is with the Hong Kong University of Science and Technology (Guangzhou), Guangzhou 511458, China. E-mail: guanjiehuang@hkust-gz.edu.cn.\protect

\IEEEcompsocthanksitem Weilin~Lin is with the Hong Kong University of Science and Technology (Guangzhou), Guangzhou 511458, China. E-mail: mlmr.lin@gmail.com.\protect

\IEEEcompsocthanksitem Yiqian~Yang is with the Northwestern Polytechnical University, Xi'an 710072, China. E-mail: frank.stuart@mail.nwpu.edu.cn.\protect

\IEEEcompsocthanksitem Yuehong~Hu is with the Central South University, Changsha 410083, China. E-mail: 8207190414@csu.edu.cn.\protect

}

\thanks{$^{*}$ Corresponding author.}
\thanks{This work was done at the Hong Kong University of Science and Technology (Guangzhou).}

}

%
%

\markboth{Journal of \LaTeX\ Class Files,~Vol.~XX, No.~XX, May~2023}%
{Shell \MakeLowercase{\textit{et al.}}: Bare Advanced Demo of IEEEtran.cls for IEEE Computer Society Journals}
%



\IEEEtitleabstractindextext{%
\begin{abstract}
Artificial intelligence (AI) is evolving towards artificial general intelligence, which refers to the ability of an AI system to perform a wide range of tasks and exhibit a level of intelligence similar to that of a human being. This is in contrast to narrow or specialized AI, which is designed to perform specific tasks with a high degree of efficiency. Therefore, it is urgent to design a general class of models, which we term foundation models, trained on broad data that can be adapted to various downstream tasks. The recently proposed segment anything model (SAM) has made significant progress in breaking the boundaries of segmentation, greatly promoting the development of foundation models for computer vision. To fully comprehend SAM, we conduct a survey study. As the first to comprehensively review the progress of segmenting anything task for vision and beyond based on the foundation model of SAM, this work focuses on its applications to various tasks and data types by discussing its historical development, recent progress, and profound impact on broad applications. We first introduce the background and terminology for foundation models including SAM, as well as state-of-the-art methods contemporaneous with SAM that are significant for segmenting anything task. Then, we analyze and summarize the advantages and limitations of SAM across various image processing applications, including software scenes, real-world scenes, and complex scenes. Importantly, many insights are drawn to guide future research to develop more versatile foundation models and improve the architecture of SAM. We also summarize massive other amazing applications of SAM in vision and beyond. Finally, we maintain a continuously updated paper list and an open-source project summary for foundation model SAM at \href{https://github.com/liliu-avril/Awesome-Segment-Anything}{\color{magenta}{here}}.

\end{abstract}

\begin{IEEEkeywords}
Survey, Artificial General Intelligence, Foundation Models, Segment Anything, Open Source Projects.
\end{IEEEkeywords}}

\maketitle

\IEEEdisplaynontitleabstractindextext

%
\IEEEpeerreviewmaketitle

\ifCLASSOPTIONcompsoc
\IEEEraisesectionheading{\section{Introduction}\label{sec:introduction}}
\else
\section{Introduction}
\label{sec:introduction}
\fi

%
%
%
%
\IEEEPARstart {F}{oundation} models~\cite{bommasani2021opportunities,wang2023large,liang2022foundations} have revolutionized artificial intelligence (AI) in the past few years, thanks to their thorough pre-training on web-scale datasets and powerful zero-shot generalization across a wide range of downstream tasks. More recently, the natural language processing (NLP) community has undergone a significant shift towards the development of large language models (LLMs), resulting in a series of ground-breaking works, \eg BERT~\cite{devlin2018bert}, T5~\cite{raffel2020exploring}, GPT-3~\cite{brown2020language}, and GPT-4~\cite{GPT4}. One of the most amazing applications of these models is the ChatGPT~\cite{ChatGPT}, an AI chatbot developed by OpenAI that leverages a large language model called GPT-3.5 to generate human-like responses to user inputs.

Due to the great success of foundation models in NLP, researchers have been inspired to explore large visual models (LVMs) in the computer vision (CV) community. One line of research is to explore scaling vision transformers to a huge size, pursuing the emergent capabilities exhibited in LLMs, such as ViT-G~\cite{zhai2022scaling}, ViT-22B~\cite{dehghani2023scaling}, Swin Transformer V2~\cite{liu2022swin}, and VideoMAE V2~\cite{wang2023videomae}. Besides, a large body of works are devoted to adding knowledge of additional modalities to enhance the capabilities of LVMs. Some notable examples include CLIP\cite{radford2021learning}, and ALIGN \cite{jia2021scaling}, which adopt a text encoder and an image encoder to learn visual and language representations of image and text pairs from massive noisy image-text data using contrastive learning~\cite{oord2018representation}. After pre-training, the learned semantic knowledge can be used to reference novel visual concepts on new data
distributions enabling the model with zero-shot transfer capability in various downstream tasks, such as image-text retrieval~\cite{plummer2015flickr30k,chen2015microsoft}, and image generation~\cite{ramesh2021zero,ramesh2022hierarchical}.

Although the progress brings new impetus to the development of CV, the generalization ability of obtained deep models remains limited~\cite{wang2023large,sam}. Recently, the CV community is witnessing a surge in exploring task-agnostic foundation models~\cite{sam,zhang2023personalize,ji2023sam,wang2023seggpt,zou2023segment,jain2022oneformer}. A common characteristic of these models is to rely on a foundation model pre-trained on a broad dataset using a task that can solve a wide range of downstream tasks using prompt learning~\cite{liu2023pre}. This new research trend of developing task-agnostic foundation models is recently sparked by a model called segment anything model (SAM)~\cite{sam} designed for general image segmentation. SAM is a promptable model trained over 1 billion masks on 11 million images using a promptable segmentation task that enables powerful zero-shot generalization. Many researchers, such as Jim Fan~\cite{GPT3moment}, consider this as ``\emph{the GPT-3 moment for CV, as SAM has learned the general concept of what an object is, even for unknown objects, unfamiliar scenes (\eg underwater and cell microscopy) and ambiguous cases}'' and demonstrated great potential as a fundamental model for CV~\cite{ji2023sam,Non-euclidean_segment,tang2023can}.

Recently, a large number of extended works have been proposed by the community to explore the capability boundaries of SAM and apply it to various tasks, \eg medical image analysis~\cite{ma2023segment,zhou2023can,shi2023generalist,zhang2023segment,wu2023medical,chen2023sam,cheng2023sam,mazurowski2023segment,huang2023segment}, image inpainting~\cite{yu2023inpaint}, image editing~\cite{xie2023edit}, style transfer~\cite{liu2023any}, infrastructure detection~\cite{ahmadi2023application}, camouflaged object detection~\cite{tang2023can}, mirror and transparent objects detection~\cite{han2023segment}, image captioning~\cite{wang2023caption}, audio-visual localization~\cite{mo2023av}, video object tracking~\cite{yang2023track}, 3D reconstruction~\cite{shen2023anything}, few-shot object counting~\cite{ma2023can}, and adversarial attacks~\cite{guan2023badsam,zhang2023attack}. Concurrent to SAM, Wang \etal~\cite{wang2023seggpt} proposed a generalist model, namely SegGPT, to unify various segmentation tasks into an in-context learning framework, which has demonstrated strong zero-shot capabilities. Furthermore, Zou \etal~\cite{zou2023segment} proposed a more general segmentation system SEEM via introducing more diverse prompts than SAM, including visual prompts (points, boxes, scribbles, masks), text prompts, and referring prompts (referred regions of another image). As the authors claim that the introduced unified prompt scheme in SEEM can encode different prompts into the joint visual-semantic space to produce a strong zero-shot generalization ability to address unseen user prompts for segmentation. Additionally, some pioneering works explore general AI methods for detecting/segmenting anything in the open-vocabulary scenarios, \eg Grounding DINO~\cite{liu2023grounding}, OVSeg~\cite{liang2022open}, V3Det~\cite{wang2023v3det}, and OpenSeg~\cite{ghiasi2021open}. These advancements have led many researchers to believe that versatile foundation models are a critical step towards artificial general intelligence (AGI)~\cite{pei2019towards,goertzel2014artificial,grudin2019chatbots,goertzel2007foundational}.

To this end, this work provides a comprehensive survey of these works with the goal of helping researchers understand the latest developments related to SAM models. This survey mainly focuses on various foundation models since SAM, especially the applications of SAM to various tasks and data types. Readers are referred to existing surveys~\cite{zhao2023survey,mialon2023augmented,fan2023bibliometric,zhang2023text,bommasani2021opportunities,wang2023large,liang2022foundations,blodgett2020language,chen2023vlp,du2022survey,long2022vision,mogadala2021trends} for language, vision, and multimodal foundation models. To the best of our knowledge, this survey is the first to comprehensively review the recent progress of segmenting anything task for vision and beyond based on the foundation model of SAM. Concurrent to our work,~\cite{zhang2023asurvey,zhang2023segment} briefly summarized recent efforts to extend SAM to vision and medical image segmentation tasks, however, we provide a more comprehensive review with many new insights from a broader perspective. Furthermore, we maintain a continuously updated paper list and a project summary to reflect the dynamic progress of the foundation model of SAM during its development.

The remainder of this survey is organized as follows: Section~\ref{sec:background_Terminology} introduces the background and terminology for foundation models including SAM, as well as methods contemporaneous with SAM that are important for segmenting anything task.
Section~\ref{sec:sam_for_image} discusses the awesome works based on SAM for various image processing applications, including software scenes, real-world scenes, and complex scenes.
Section~\ref{sec:other_applications} further discusses the follow-up works of SAM that extend SAM to vision-related applications, beyond vision, and in more directions. Finally, we conclude the survey in Section~\ref{sec:conclusion}. This survey will be regularly updated to reflect the dynamic progress of the foundation model of SAM, as this is a rapid-evolving and promising field towards AGI.

\section{Background and Terminology}
\label{sec:background_Terminology}
\subsection{Image Segmentation}
\subsubsection{Classic Segmentation}
Image segmentation is a fundamental computer vision task that separates a digital image into multiple parts by assigning each pixel to a class or object. Traditionally, the segmentation includes three major tasks: semantic, instance, and panoptic. Semantic segmentation \cite{chen2014semantic,chen2017rethinking,chen2017deeplab,chen2018encoder} assigns each pixel to a predefined semantic class label. Instance segmentation \cite{hafiz2020survey,liu2018path,bolya2019yolact} further separate instances of the same class. Panoptic segmentation proposed by \cite{Kirillov_2019_CVPR} combines semantic and instance segmentation to understand scenes comprehensively. Researchers have fully explored the above tasks in past studies. Due to the operation consistency of the above tasks at the pixel level, many studies have tried to use a unified framework to provide solutions for three segmentation methods simultaneously, such as K-net \cite{zhang2021knet}, MaskFormer\cite{cheng2021maskformer}, and Mask2Former\cite{cheng2022maskedattention}.

\subsubsection{Interactive Segmentation}
Interactive segmentation \cite{xu2016deep} is a particular segment task with the character of leveraging information from the guidance of user interaction. Despite being a longstanding challenge, the problem has seen considerable improvement. Usually, the user provides some initial input, such as points, strokes, or bounding boxes, to indicate the rough location and shape of the object. Then, the algorithm iteratively refines the segmentation based on the user feedback, such as correcting mislabeled regions or adding missing parts. Interactive segmentation is useful for many applications that require precise object extraction, such as medical image analysis \cite{Wang_2019}, \cite{mohan2018mri}, \cite{yushkevich2016itk}, photo editing \cite{rupprecht2018guide}, and data annotation \cite{castrejon2017annotating}, \cite{acuna2018efficient}.

\begin{figure*}[t]
    \centering
    \includegraphics[width=0.9\textwidth]{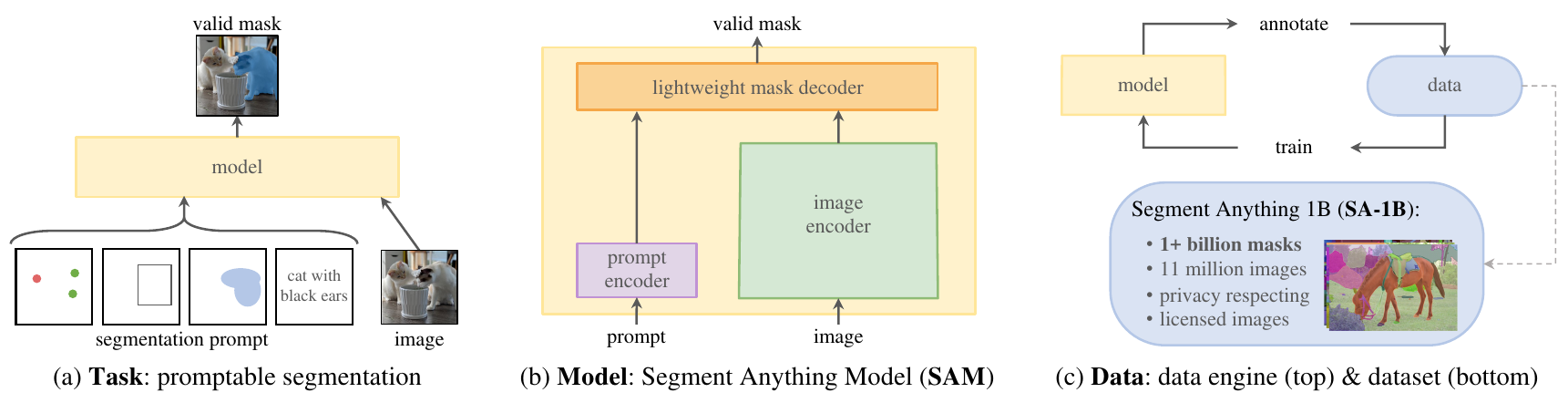}
    
    \caption{Overview of the SA project, including task, model, and data. The figure is borrowed from the original paper~\cite{sam}.}
    \label{fig:SA_structure}
\end{figure*}

\begin{figure*}[htbp]
    \centering
    \includegraphics[width=0.9\textwidth]{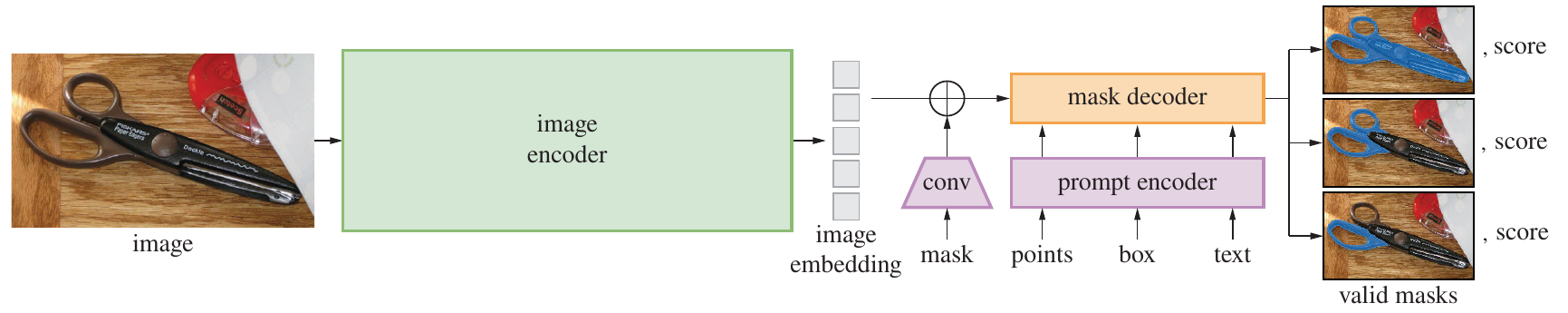}
    \caption{Overall structure of SAM from the original paper~\cite{sam}.}
    \label{fig:SAM_structure}
\end{figure*}

\subsection{Foundation Models}
Foundation models are a new paradigm for building artificial intelligence systems that can be adapted to various downstream tasks. They are based on training large neural networks on massive amounts of data, often using self-supervised learning techniques. This allows them to learn general representations and capabilities that can be transferred to different domains and applications. The term was coined by the Stanford Center for Research on Foundation Models (CRFM) in 2021 to capture the significance and challenges of this paradigm \cite{bommasani2021opportunities}.

The development of foundation models can be traced back to the rise of deep learning and self-supervised learning in the NLP field, which enabled learning powerful representations from raw text data. Early examples of foundation models were pre-trained LLMs, such as BERT \cite{devlin2018bert}, T5 \cite{raffel2020exploring}, and GPT-n series \cite{brown2020language}, \cite{ChatGPT}, \cite{GPT4}, which demonstrated impressive capabilities and performance on a wide range of NLP tasks.

In CV research, current foundation models try to take advantage of LLMs which are trained on large-scale data and show superb performance in learning universal visual representations from diverse, large-scale image-text data. Representatives include CLIP \cite{radford2021learning}, ALIGN \cite{jia2021scaling}, Florence \cite{yuan2021florence}, VLBERT \cite{su2020vlbert}, X-LXMERT \cite{cho2020xlxmert}, and DALL-E \cite{ramesh2022hierarchical} try to capture the cross-modal interactions between vision and language. They can be transferred or direct act on classification, retrieval, object detection, video understanding, visual question-answering, image captioning, and image generation tasks.
Recently, ImageBind~\cite{girdhar2023imagebind} attempted to align six different modal information around image/video information and learn the unified embedding space, opening up further research on multimodal foundation models. Foundation models for computer vision and multimodal learning are still an active area of research, with many challenges and opportunities for improving their performance, robustness, interpretability, and social impact.

\subsection{Segment Anything Model}

SAM comes from the Segment Anything (SA) project of Meta in 2023 \cite{sam}. By finding foundation models that appeared in the NLP and CV fields show strong performance, researchers tried to build a similar model which can unify the whole image segmentation task. However, the available data in the segmentation field is insufficient and differs from their design purpose. Therefore, as shown in Fig. \ref{fig:SA_structure}, they divide the pathway into three steps, namely Task, Model, and Data. Correspondingly, a project for segmentation tasks is proposed, including the promptable segmentation task (prompts include providing a location, a range, a mask, or a text description of the segmentation target), the SAM that can accept multiple prompt inputs and realize interactive use and the Dataset SA-1B formed using the data engine of the interactive train-annotate loop process with over one billion masks.

\subsubsection{Task}
The ultimate goal of the SA project is to provide a model with a wide range of functions that can be quickly adapted to many existing and new segmentation tasks (such as performing edge detection, object proposal generation, instance segmentation, and segmenting objects from free-form text), and can transfer zero samples to new data distributions and tasks. Since many complex functions can be realized through simple combinations of existing tools. For example, if there is a bounding box detector for humans, human instance segmentation can be solved by providing the box output of the detector as a prompt to the model. Researchers take inspiration from LLMs to achieve this target, using prompt engineering \cite{brown2020language} to cover the pre-training and downstream tasks. More specifically, The concept of interactive segmentation is introduced to form the promptable task and realize the training of the model.

A unique characteristic of the promptable task is returning a valid segmentation mask when given any segmentation prompt. A prompt can be anything indicating what to segment. A valid segmentation mask means that even if the input prompt would lead to ambiguity (such as an image of a human wearing a T-shirt, the prompt point is on the T-shirt), it should be a reasonable mask for at least one object (which returns the mask of the human or The masks of the T-shirt are all reasonable).

\subsubsection{Model}

 The structure of SAM is shown in \ref{fig:SAM_structure}. It mainly consists of three parts, a powerful image encoder (MAE \cite{he2021masked} pre-trained ViT \cite{dosovitskiy2021image}); a prompt encoder, which is divided into sparse input (CLIP's \cite{radford2021learning} text encoder is used as a position encoder to process points, boxes, and text-form prompt) and dense input (convolutions processes mask input); and a mask decoder (prompt-image bidirectional Transformer decoder using self-attention and cross-attention). In addition, when the input prompts are ambiguous, the network will rank the three possible mask outputs based on confidence. The loss functions used in training include focal loss \cite{lin2018focal} and dice loss \cite{milletari2016vnet}.
\subsubsection{Data}
Since there is insufficient public data for training, researchers use the training-annotation iterative process to form a Data Engine to achieve model training and dataset construction simultaneously. The specific process can be divided into three stages.
\begin{enumerate}

    \item Assisted-manual stage. Professional annotators use the interactive labeling tool on the browser, combined with SAM for manual labeling. SAM first uses the public dataset for training. As the data gradually increases, the size of the image encoder of SAM also increases. At the end of this stage, 4.3M masks and 120k images were collected.
    \item Semi-automatic stage. To increase the mask's diversity and improve the model's performance, the researchers first pre-filled the mask with which the model can make high-confidence predictions. Then they asked the annotator to annotate the unfilled part interactively. At the end of this stage, an image can provide an average of 72 masks.
    \item Fully automatic stage. In this stage, due to the collection of enough masks and the introduction of the ambiguity-aware model, the final training of the SAM and the acquisition of the SA-1B data set can be performed. The ambiguity-aware model enables SAM to predict effective masks even when the prompt is ambiguous. Specifically, the researchers use a 32x32 grid to obtain prompt points on each image uniformly. If the prompt point is located on the target part or sub-part structure, the model will return the mask of the sub-part, part, or the whole object. And filter sort the outputs based on confidence. At the end of this stage, the final SA-1B dataset contains 11M images and 1.1B masks.
\end{enumerate}

With the advantages of well-designed tasks, model structure, and massive high-quality training data, experimental shows that the zero-sample transfer capability of the SAM model has performed excellent results in single cue point segmentation, edge detection, Object proposal, instance segmentation, interactive segmentation, and multimodal segmentation (Text-to-Mask) tasks. It even outperforms supervised models in some respects.

\subsection{Concurrent Works}
Parallel to SAM research, many efforts have been made to solve segmentation tasks with other general methods.

OneFormer\cite{jain2022oneformer} leverages the task-conditioned joint training strategy, task token, and query-text contrastive loss to form a universal image segmentation framework. Oneformer enables training on all three traditional segmentation tasks within a single universal model and a multi-task training process. With different backbones, it outperforms specialized models on  \cite{zhou2018semantic}, Cityscapes \cite{Cordts2016Cityscapes}, and COCO datasets \cite{lin2015microsoft} with even costs much less training time and resources.

Meanwhile, the SegGPT\cite{wang2023seggpt} stand for \textbf{Seg}ment everything with a \textbf{G}eneralist \textbf{P}ain\textbf{t}er, exploring in-context training and inference scheme. It forms a generalist in-context learning framework \cite{wang2023images} that unifies different segmentation data formats. And treating the training process as an in-context coloring problem with a random coloring scheme instead of using predefined color space. This training process requires the model to focus on contextual information to accomplish the specific task. Based on these improvements, the model can perform arbitrary segmentation tasks based on input images or videos through in-context inference.

Also, SEEM\cite{zou2023segment} further broadens the scope of task applicability of single segmentation models. It further expands the types of supported prompts, including points, boxes, scribbles, masks, texts, and referred regions of another image. With the proposed joint visual-semantic space, the model has compatibility to compose flexible multi-prompt input. SEEM also can process as a classic segmentation model when no prompt provides. However, it also suffers from limited training data and absent the support of part-based segmentation.

\section{SAM for Image Processing}
\label{sec:sam_for_image}
\subsection{Software Scenes}
\subsubsection{Image Editing}

Modern software scenes require operations on image editing and inpainting, \eg removing objects, filling objects, and replacing objects. However, the existing inpainting works, like~\cite{suvorov2022resolution, lugmayr2022repaint, li2022mat, dong2022incremental}, need fine annotations for each mask to achieve good performance, which is labor-intensitive. SAM~\cite{sam}, which can generate accurate masks with simple prompts such as points or boxes, can help assist the image editing scenes.

\begin{figure}[htbp]
    \centering
    \includegraphics[width=0.5\textwidth]{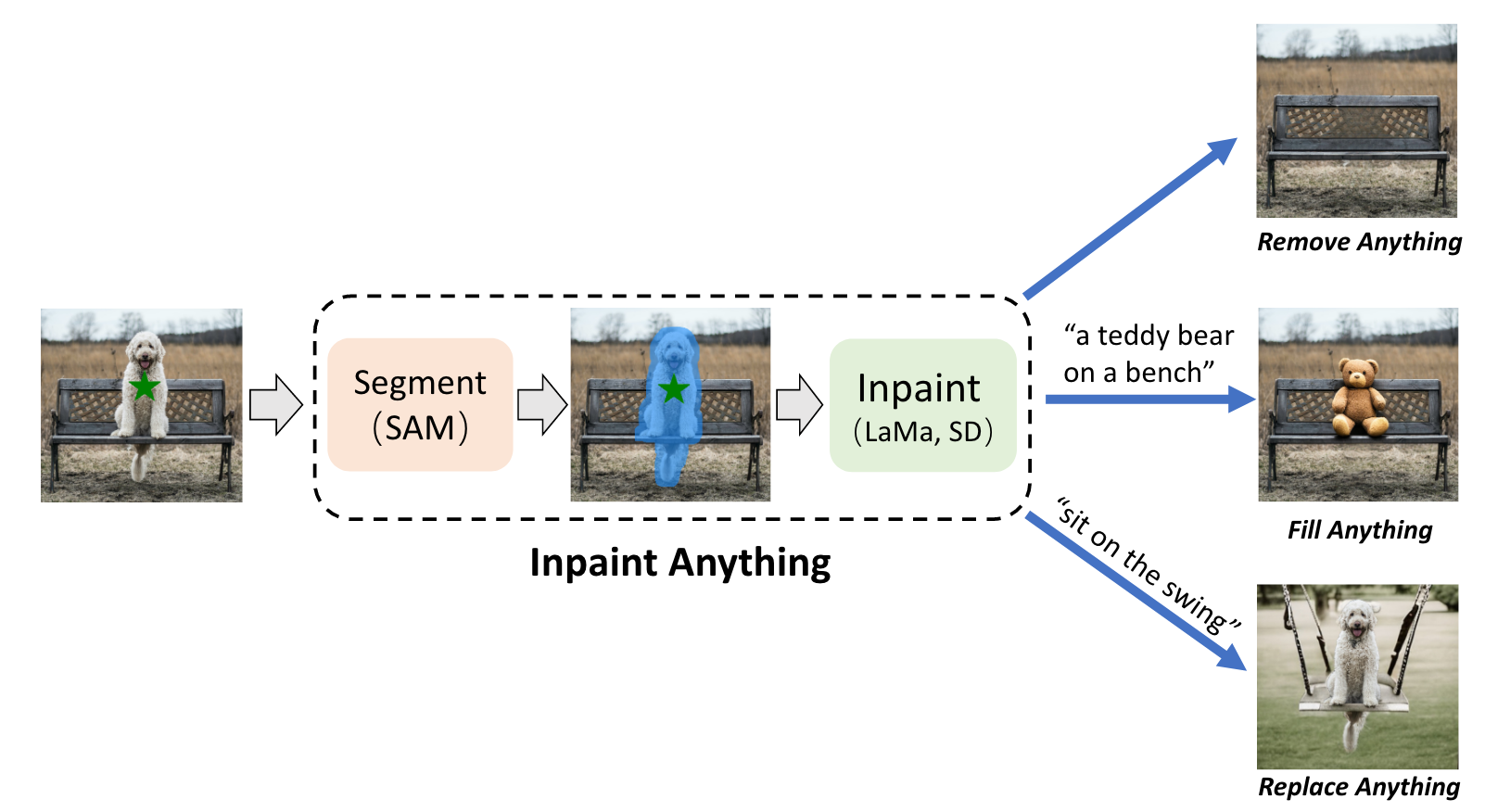}
    \caption{Overall pipeline of Inpaint Anything (IA). The input image is segmented by the SAM and the targeted segment is replaced by the output of the inpaint models to achieve different tasks. The figure is borrowed from the original paper~\cite{yu2023inpaint}.}
    \label{fig:ip_inpaintAnything}
\end{figure}

Inpaint Anything (IA)~\cite{yu2023inpaint} designs a pipeline to solve inpainting-related problems by combining the advantages of SAM, the state-of-the-art (SOTA) image inpainters~\cite{suvorov2022resolution}, and AI-generated content (AIGC) models~\cite{rombach2022high}. The pipeline is illustrated in Fig.~\ref{fig:ip_inpaintAnything}. For object removing, the pipeline is composed of SAM and SOTA inpainters, like LaMa~\cite{suvorov2022resolution}. The clicking action from the user is used as a prompt in SAM to generate a mask for the object area, and the LaMa fills it with corrosion and dilation operations. For object filling and replacing, the AIGC models, like Stable Diffusion (SD)~\cite{rombach2022high}, are used in the second step to fill the selected object with newly generated objects by text prompts.

\begin{figure}[htbp]
    \centering
    \includegraphics[width=0.495\textwidth]{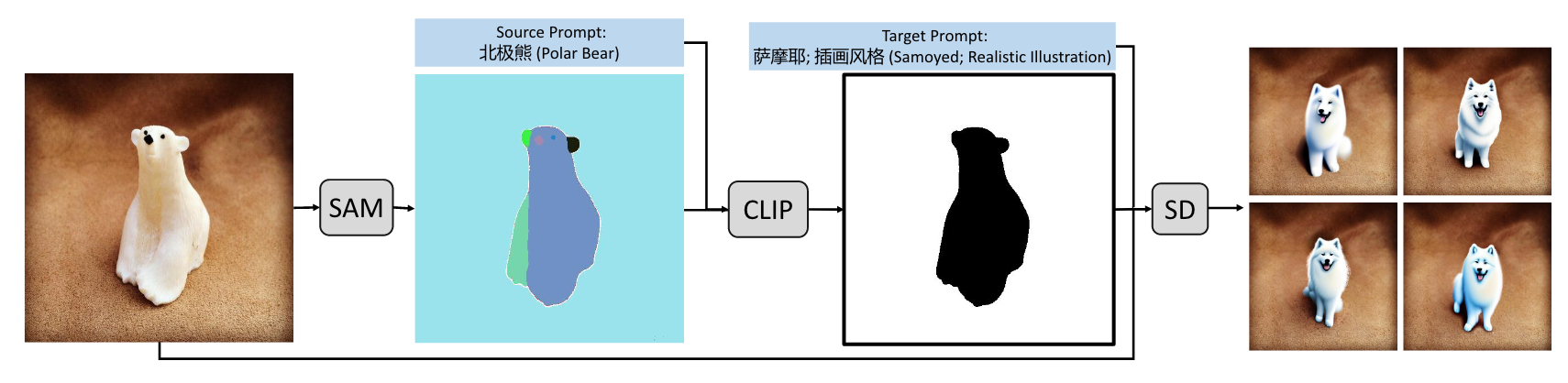}
    \caption{Overall pipeline of Edit Everything from the original paper~\cite{xie2023edit}.}
    \label{fig:ip_editEverything}
\end{figure}

A similar idea can also be seen on Edit Everything~\cite{xie2023edit}. As shown in Fig.~\ref{fig:ip_editEverything}, it allows users to edit images using simple text instructions. Specifically, when inputting an image, SAM first separates it into several segments with no prompts, followed by a source prompt instructing CLIP to rank the received segments. Only the segment with the highest score is selected as the target one to be replaced with the newly generated object from SD with the target prompt. Compared with the object-replacing solution in IA, the authors train the CLIP with 400 million parameters and the SD with 1 billion parameters in Chinese scenarios to make it more reliable towards the Chinese text prompts. Furthermore, the paper improves the realism of the image by breaking down the complex prompts into smaller entities and replaced in a sequential manner. Although it performs well as a novel tool, the paper points out that it still needs specific enhancement in different scenarios.

\subsubsection{Style Transfer}
\begin{figure}[htbp]
    \centering
    \includegraphics[width=0.5\textwidth]{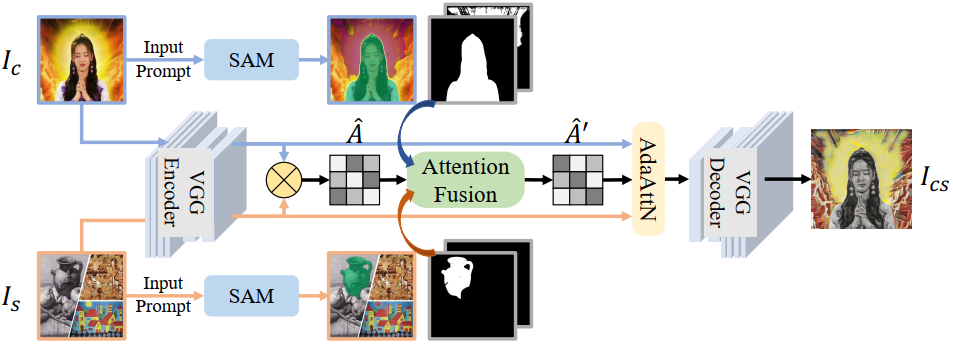}
\caption{Illustrations of Any-to-Any Style Transfer from the original paper~\cite{liu2023any}.}
    \label{fig:ip_styleTransfer}
\end{figure}
Style transfer aims to transfer the style from a given image (\textit{style image}) to another given image (\textit{content image}). Typically, the transferred style is represented by the holistic style of the style image or the local colors and textures of the style image, and only one result will be generated for the content image, which lacks flexibility for the users to interact with it. With promptable region selection ability from SAM, Any-to-Any Style Transfer~\cite{liu2023any} enables users to specify which style region to select and which content regions to apply during style transfer. The pipeline, which is shown in Fig.~\ref{fig:ip_styleTransfer}, is organized as follows:
\begin{itemize}
\item [1)] Encode the style and content images with pre-trained VGG-19 as well as calculate the content-style attention map.
\item [2)] Obtain the style and content masks with SAM and input prompts.
\item [3)] Fuse the attention map with the mask-controlling signals from the last step.
\item [4)] Compute the stylized feature with the updated attention map and derive the final result.
\end{itemize}

With the defined pipeline, the paper proves that the proposed method is a plug-and-play component to the existing style transfer methods, including Local Transformation Based Style Transfer~\cite{chen2016fast, park2019arbitrary}, Global Transformation Based Style Transfer~\cite{li2017universal}, and Diffusion Based Style Transfer~\cite{xu2023stylerdalle}, which shows its great potential of broad applications.

\subsection{Real-World Scenes}
\subsubsection{Detection}
SAM holds the ability to assist the applications in many real-world scenes such as real-world object detection, object counting, and moving object detection scenes. Recently, \cite{ji2023segment} evaluated the performance of SAM across a diverse range of real-world segmentation scenarios, \eg natural image, agriculture, manufacture, remote sensing, and healthcare scenes. The paper finds that it has excellent generalization on common scenes like natural images, while it shows less effectiveness in low-contrast scenes and requires strong prior knowledge in complex scenes. 

\begin{figure}[htbp]
    \centering
    \includegraphics[width=0.5\textwidth]{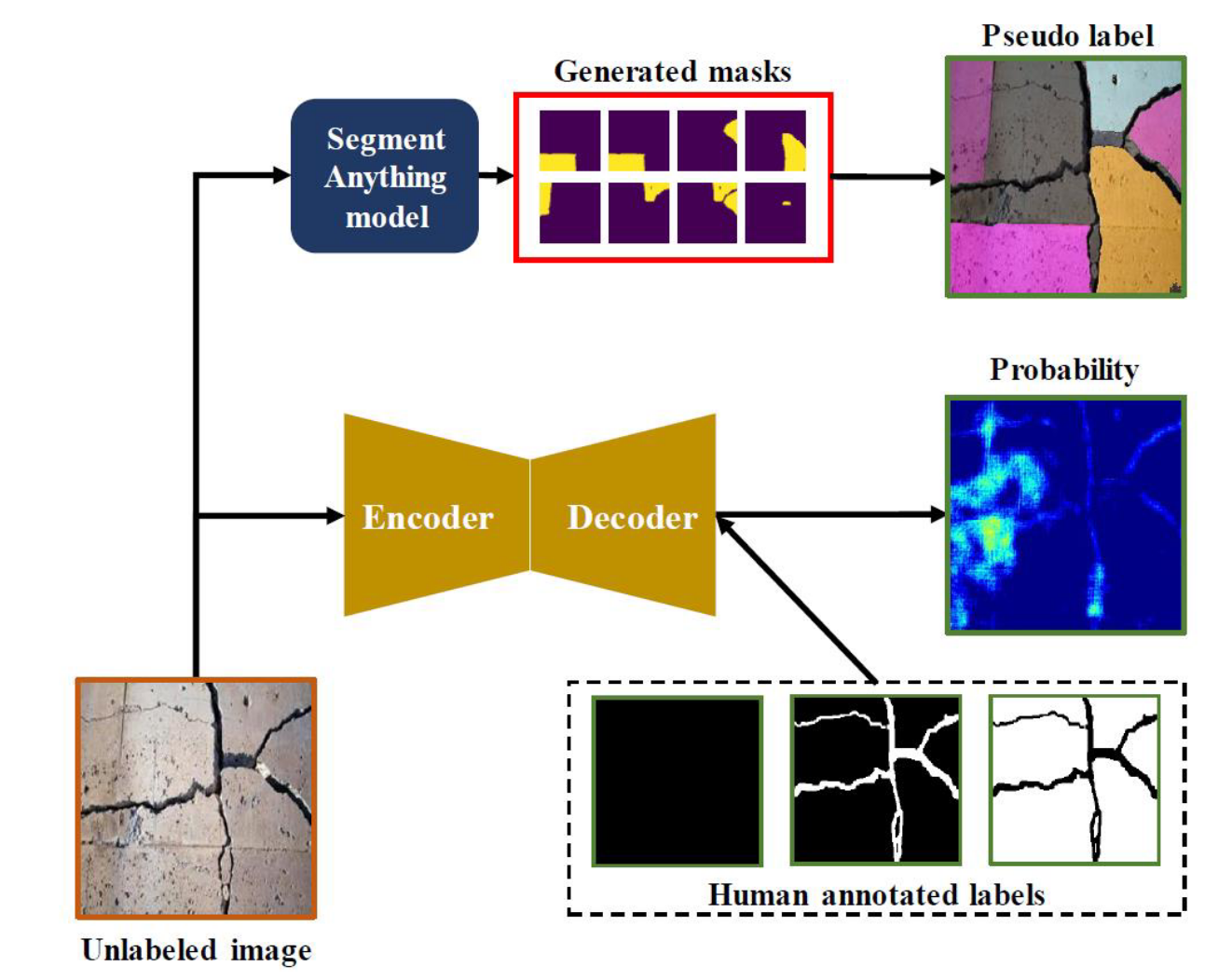}
\caption{Process of crack detection using SAM and U-Net. The figure is borrowed from the original paper~\cite{ahmadi2023application}.}
    \label{fig:ip_civilDetect}
\end{figure}

For example, in the application of civil infrastructure defect assessment, \cite{ahmadi2023application} utilizes SAM to detect cracks in concrete structures and compares its performance with the baseline U-Net~\cite{ronneberger2015u}. The crack detection process is illustrated in Fig.~\ref{fig:ip_civilDetect}. The results showed that the SAM outperforms the U-Net in the detection of longitudinal cracks, which are more likely to find similar training images as in the normal scenes, while in the unusual scene, \ie spalling cracks, SAM is not as good as U-Net.

Unlike the complex image cases in crack detection, crater detection is more suitable to utilize SAM as the detection tool as the crater shapes focus on circular or elliptical. Craters are one of the most important morphological features in planetary exploration, detecting and counting them is an important but time-consuming task in planetary science. Although the existing works in machine learning and computer vision successfully solve some specific problems in crater detection, they rely on specific types of data and thus fail to work well in a different data source. 

In \cite{giannakis2023deep}, the authors propose a universal crater detection scheme with the zero-shot generalization of SAM to unfamiliar objects. The pipeline uses SAM to segment the input images, which has no restrictions on the data type and resolutions. Then, it utilizes circular-elliptical indexes to filter the segmentation masks that are not circular-elliptical shapes. Finally, a post-processing filter is employed to get rid of duplicates, artifacts, and false positive ones. The pipeline shows its great potential to be the general tool in the current field and the authors also discuss the drawbacks that only the specific shapes can be recognized.

\subsubsection{Counting}
Few-shot object counting is an important application scene of computer vision in the real world, which counts an unseen-category object with only a few bounding boxes of examples provided. Since SAM shows impressive performance with great generalization for unseen objects, it shows potential to be used in few-shot object counting. \cite{ma2023can} is the first paper testing SAM in this task and comparing it with other baseline few-shot counting methods. The goal is to find out whether SAM can segment and distinguish the target objects using the reference examples. 

To realize this, the authors define a pipeline. Firstly, they calculate the dense image feature with an image encoder, \ie ViT-H. Secondly, they use bounding boxes as prompts to generate segment masks for the reference examples, which are then computed with the dense image feature as the feature vectors of the reference objects. Thirdly, they use point grids as prompts to segment everything and generate feature vectors for all masks. After that, they compute the cosine similarity of the feature vectors between the predicted masks and reference examples, only the ones larger than the pre-defined threshold are considered the target objects. The paper compares the 
proposed SAM-based method with other few-shot counting methods on two datasets, FSC-147~\cite{ranjan2021learning} and MS-COCO~\cite{lin2014microsoft}, and finds that it falls behind the SOTA baselines, especially for small and congested objects. Thus, further improvement for SAM in some special scenes is still needed.

\subsubsection{Moving Object}
Moving object segmentation (MOS) is a crucial task for the application of computer vision in many real-world application scenarios, such as autonomous driving. The existing datasets for this research are mainly RGB or Lidar videos, which lack the event information that can help understand the dynamic scenes better. To fill this gap, DSEC-MOS~\cite{zhou2023dsec} is proposed with the moving vehicles' frames and the corresponding event data, which can facilitate the development of more accurate, robust, and efficient algorithms for autonomous driving. The contribution of SAM to DSEC-MOS annotation is that it provides a promptable segmentation way. The authors apply the moving object bounding boxes in DSEC-MOD~\cite{zhou2022rgb} as prompts to generate a large number of preparatory masks with SAM, which is accurate and reliable. The dataset contains 16 sequences of 13,314 frames in total and provides event-based data with pixel-level annotations, which can be a valuable resource for the MOS field.

\begin{figure}[htbp]
    \centering
    \includegraphics[width=0.99\linewidth]{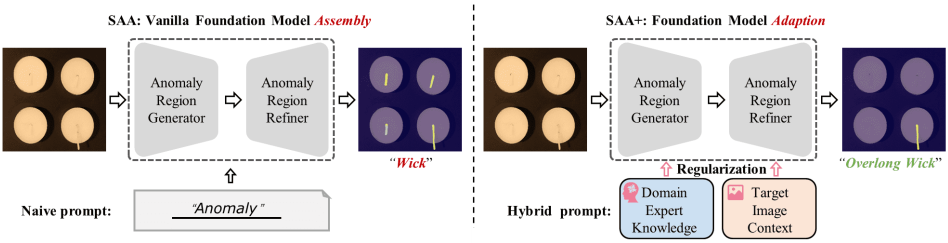}
\caption{Framework of the SAA+~\cite{Yunkang2023SAA}, which achieves zero-shot anomaly segmentation by introducting hybrid prompts as a regularization technique. The figure is borrowed from the original paper~\cite{Yunkang2023SAA}.}
    \label{fig:SAA+}
\end{figure}

\subsection{Complex Scenes}
\subsubsection{Low-Contrast Scene}
Except for the normal scenes mentioned above, whether SAM can solve the segmentation issue in complex scenes, like low-contrast scenes, is also a meaningful question to broaden its application. To find out the generalization ability of SAM in more complex scenes, Ji \etal~\cite{ji2023sam} quantitatively compare it with cutting-edge models in three concealed scenes, \ie camouflaged animals, industrial defects, and medical lesions. They conduct the experiments on three camouflaged object segmentation (COS) datasets, \ie CAMO~\cite{le2019anabranch} with 250 samples, COD10K~\cite{fan2020camouflaged} with 2026 samples, and NC4K~\cite{lv2021simultaneously} with 4121 samples. And compare it with the outstanding transformer-based model, CamoFormer-P/S~\cite{yin2022camoformer} and HitNet~\cite{hu2022high}. They observe from the results that SAM looks not skillful in concealed scenes and point out that the potential solution may rely on the support of prior knowledge in the specific fields. The same conclusion also can be drawn in \cite{tang2023can}, where the authors compare SAM with 22 SOTA methods in camouflaged object detection on the same three datasets mentioned above. 

Dominic \etal~\cite{williams2023leaf} utilizes SAM in the field of plant phenotyping, \eg potato leaves segmentation, by proposing a method named Leaf Only SAM. Specifically, they combine SAM with four post-processing steps to identify only leaf objects without any training data. After receiving the segmentation masks from SAM, they first check the color by finding the green masks. To reduce ambiguity, they further check the main object by keeping the one with IoU of more than 90\% and removing the rest. Then, they compare the area of the minimum enclosing circle to filter the masks with incorrect shapes. Finally, they remove multi-leaf masks by summing all the mask objects in the image and labeling each pixel to the amount of belonging masks. The mask with a mean score over 1.5 is assumed to be a duplicate. The experiments compare Leaf Only SAM with a fine-tuned Mask R-CNN and find that Leaf Only SAM does not outperform the latter with just a 10\% performance drop, but its training-free ability shows its potential for this field.

\begin{figure}[htbp]
    \centering
    \includegraphics[width=0.495\textwidth]{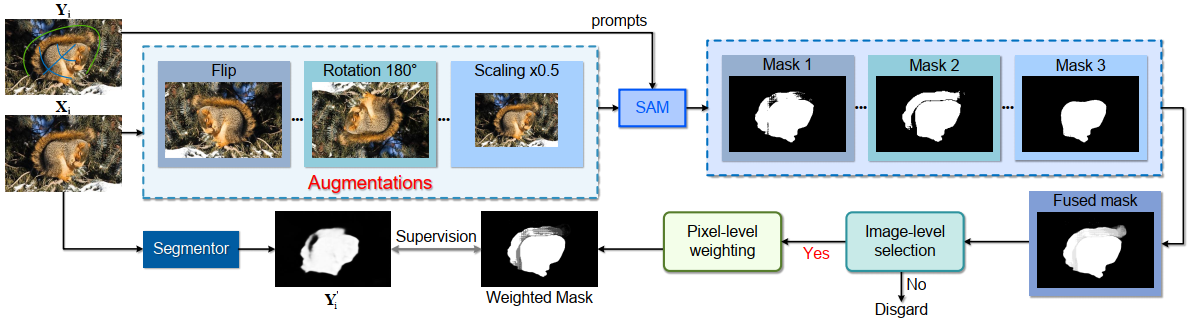}
\caption{Framework of the WS-SAM~\cite{Chunming2023WSCOS} with scribble supervision for weakly-supervised concealed object segmentation. The figure is borrowed from the original paper~\cite{Chunming2023WSCOS}.}
    \label{fig:wsSAM}
\end{figure}

Cao \etal~\cite{Yunkang2023SAA} proposes a new framework called Segment Any Anomaly + (SAA+) for zero-shot anomaly segmentation as shown in Fig.~\ref{fig:SAA+}. The framework utilizes hybrid prompt regularization to improve the adaptability of modern foundation models, allowing for more accurate anomaly segmentation without the need for domain-specific fine-tuning. The authors conduct thorough experiments on four anomaly segmentation benchmarks, \ie VisA~\cite{zou2022spot}, MVTec-AD~\cite{bergmann2019mvtec}, MTD~\cite{huang2020surface}, and KSDD2~\cite{bovzivc2021mixed}, and achieve SOTA performance. He \etal~\cite{Chunming2023WSCOS} proposes the first method (WS-SAM) to leverage SAM for weakly-supervised concealed object segmentation that addresses the challenges of segmenting objects that are well blended with their surroundings using sparsely-annotated data (see Fig.~\ref{fig:wsSAM}). The proposed WS-SAM involves SAM-based pseudo labeling and multi-scale feature grouping to improve model learning and distinguish concealed objects from the background. The authors found that using only scribble supervision~\cite{gao2022weakly}, SAM can generate segmentation masks good enough to train a segmenter.

The glass scene is a more challenging one in concealed scenes compared with the camouflaged animals and others. It is closely related to the safety issue when applying the machine learning algorithm in our real world. For example, autonomous mobile robots may easily crash into a transparent front door without a reliable, transparent object detection algorithm. In this scene, the detected objects often have mirror surfaces or just be transparent, which easily makes the traditional detection algorithms fail. As SAM has strong capability in zero-shot segmentation to the unseen objects, \cite{han2023segment} evaluates it on both mirror and transparent objects scenes and compares it with SOTA methods from semantic segmentation, shadow detection, saliency object detection, and glass segmentation. The results from the experiments show that SAM can successfully identify the objects behind the transparent ones but fail to recognize the glass objects themselves. The overall performance of SAM is significantly worse than the methods trained specifically with transparent objects, which proves that SAM is not ready to be deployed in safety-critical situations that contain glass.

\subsubsection{Thermal Infrared Image}
\begin{figure}[htbp]
    \centering
    \includegraphics[width=0.5\textwidth]{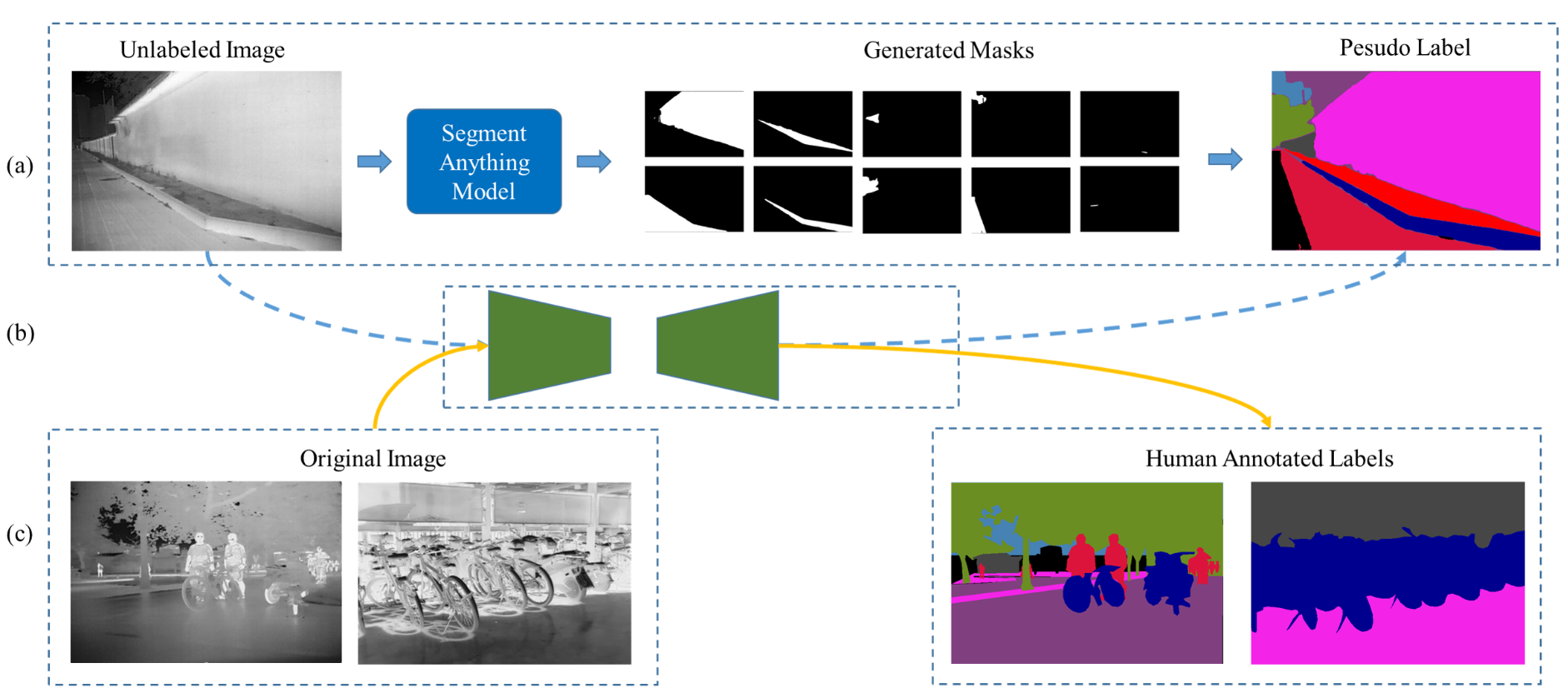}
    \caption{Framework of creating and using SATIR in three steps. (a) Construct a pre-trained dataset with SAM. (b) Pretrain models with the dataset. (c) Finetune the pre-trained models on the target mask. The figure is borrowed from the original paper~\cite{chen2023learning}.}
    \label{fig:ip_thermal}
\end{figure}
The thermal infrared image scene is another kind of complex scene where the images are always dark and hard to be annotated. Therefore, a large amount of unlabeled data is wasted, and the models for this field fail to learn high accuracy in a reliable way. To solve this problem, \cite{chen2023learning} utilizes SAM to generate pseudo labels and builds a large-scale thermal infrared segmentation dataset, SATIR, for model pretraining, which contains over 100,000 images with pixel-annotation labels. To finally improve the performance of models in this field, the authors propose a three-step framework, which is shown in Fig.~\ref{fig:ip_thermal}. They construct the mentioned dataset with SAM, followed by pretraining the model with it. Then, they finetune the pre-trained model on the target task. The experiment on the public thermal infrared semantic segmentation data, SODA~\cite{li2020segmenting}, validates its effectiveness in this field, where the backbone model pre-trained by SATIR outperforms others with approximately 1.3\% in mean Intersection over Union (mIoU). 

In the field of the poultry industry, thermal infrared images are also used in chicken segmentation tasks, which is an important agricultural application in the context of cage-free hens. To find out SAM's capacity in this field, the paper~\cite{yang2023sam} uses SAM in the chicken segmentation task and compares its performance with the other SOTA baseline methods, \ie SegFormer~\cite{xie2021segformer} and SETR~\cite{liu2022setr}. The authors find that SAM has superior performance compared with the two baseline methods, especially when using the total points prompts. However, with the thermal images as input, they find that there are still some limitations in recognizing the arbitrary parts, \eg the chicken tail. Except for the chicken segmentation, they also explore the ability of SAM in chicken-tracking tasks by combining SAM, YOLOX~\cite{ge2021yolox}, and ByteTracker~\cite{zhang2022bytetrack} as a tracker. And they realize real-time tracking with the combination and show its potential in this field.

\subsubsection{Overhead Image}
Overhead imagery problems are related to a set of widely-studied tasks, where the objects are usually small and intensive, \eg the remote sensing images. In \cite{ren2023segment}, the authors test whether the impressive 
generalization ability of SAM can cover the scenes in this field. Specifically, they examine the out-of-the-box accuracy of SAM on six existing benchmark datasets of overhead imagery, including Solar~\cite{bradbury2016distributed}, Inria~\cite{maggiori2017can}, DeepGlobe~\cite{demir2018deepglobe}, 38-Cloud~\cite{mohajerani2018cloud}, DeepGlobe Roads~\cite{demir2018deepglobe}, and Parcel Delineation~\cite{aung2020farm}, which encompass 5 million square kilometers of surface area. The results demonstrate that SAM often generalizes well to overhead imagery while failing in some cases with unique characteristics on target objects, which is considered a systemic failure and can be addressed through modest changes in SAM's design. Similar observations and ideas are proposed in \cite{julka2023knowledge}. The authors adopt SAM to the geological mapping task with remote sensing images of the Mars surface and find that it cannot be directly applied to domain-specific tasks due to a lack of problem-specific bias. Thus, they change the SAM's design by introducing a domain-specific decoder, which can learn the problem-specific semantics by fine-tuning through knowledge distillation with only five labeled images and achieve the same performance as the Mask-RCNN~\cite{he2017mask} trained with over 400 labeled images.

The unlabeled remote sensing dataset issue can also be solved with SAM. With the small and intensive characters on objects in remote sensing images, they are hard and cost-inefficient to be labeled by human experts. Thus, the paper~\cite{wang2023scaling} develops an efficient pipeline with SAM to generate a large-scale remote sensing segmentation dataset named SAMRS. It will be detailed in the data annotations section. The paper~\cite{julka2023knowledge} mentioned above also explores the applicability of their proposed SAMs decoder for annotation and finds that the bounding box-based segmentation mode is more suitable for rapid annotation. Apart from that, some researchers combine the advantages of different foundation models~\cite{text2seg}, like SAM and Grounding DINO, to achieve text-prompt guiding segmentation on remote sensing images and prove its effectiveness in this field. The details are illustrated in the vision and language section.

\section{Other Applications: Vision and Beyond}
\label{sec:other_applications}
\subsection{Vision Related Applications}


\begin{figure*}[t]
   \vspace{-0.05in}
   \begin{overpic}[width=1\textwidth]{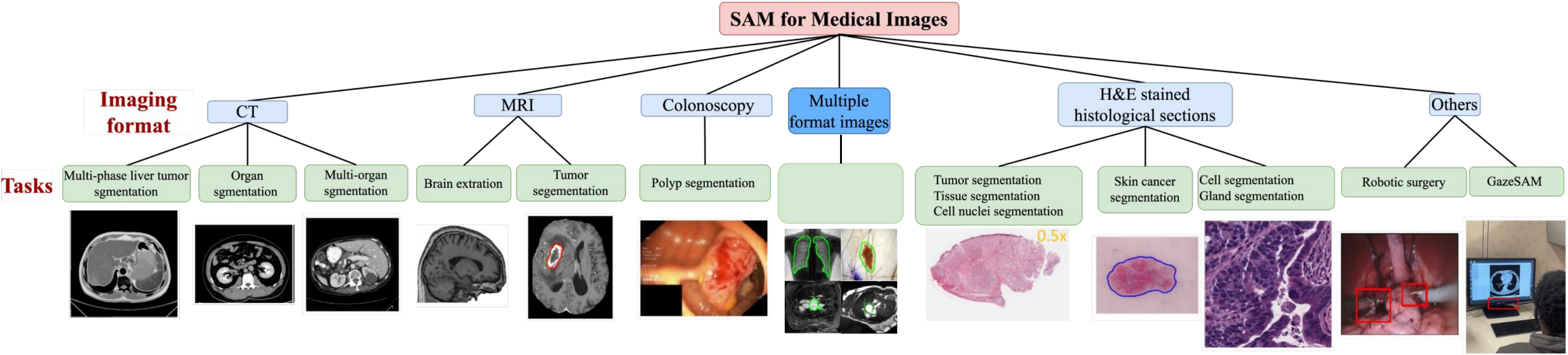}
   \put(10.0,10.14){\fontsize{3.5pt}{\baselineskip}\selectfont \color{black}{\cite{hu2023sam}}}
   \put(16.8,11.15){\fontsize{3.5pt}{\baselineskip}\selectfont \color{black}{\cite{roy2023sam}}}
   \put(24.4,10.7){\fontsize{3.5pt}{\baselineskip}\selectfont \color{black}{\cite{zhang2023customized}}}
   \put(28.75,10.0){\fontsize{3.5pt}{\baselineskip}\selectfont \color{black}{\cite{mohapatra2023brain}}}
   \put(37.5,11.25){\fontsize{3.5pt}{\baselineskip}\selectfont \color{black}{\cite{putz2023segment}}}
   \put(43.3,9.95){\fontsize{3.5pt}{\baselineskip}\selectfont \color{black}{\cite{li2023polyp,zhou2023can}}}
   \put(49.8,11.0){\fontsize{3.5pt}{\baselineskip}\selectfont \color{black}{\cite{ma2023segment,mazurowski2023segment,he2023accuracy,shi2023generalist,cheng2023sam}}}
   \put(49.8,10.0){\fontsize{3.5pt}{\baselineskip}\selectfont \color{black}{\cite{mattjie2023exploring,huang2023segment,liu2023samm,wu2023medical,chen2023sam}}}
    \put(49.8,9.0){\fontsize{3.5pt}{\baselineskip}\selectfont \color{black}{\cite{qiu2023learnable,junde2023PromptUNet}}}
   
   \put(66.3,9.9){\fontsize{3.5pt}{\baselineskip}\selectfont \color{black}{\cite{deng2023segment}}}
   \put(72.3,9.15){\fontsize{3.5pt}{\baselineskip}\selectfont \color{black}{\cite{hu2023skinsam}}}
   \put(82.5,10.9){\fontsize{3.5pt}{\baselineskip}\selectfont \color{black}{\cite{zhang2023input}}}
   \put(88.7,10.15){\fontsize{3.5pt}{\baselineskip}\selectfont \color{black}{\cite{wang2023sam}}}
   \put(98.15,10.88){\fontsize{3.5pt}{\baselineskip}\selectfont \color{black}{\cite{wang2023gazesam}}}
   \end{overpic}
   \caption{Outline of SAM for medical images, including Computerized Tomography (CT) images, Magnetic Resonance Imaging (MRI) images, colonoscopy images, multiple format images, H\&E stained histological sections images, and others.} 
   \label{fig:medicalimage}
   \vspace{-0.05in}
\end{figure*}
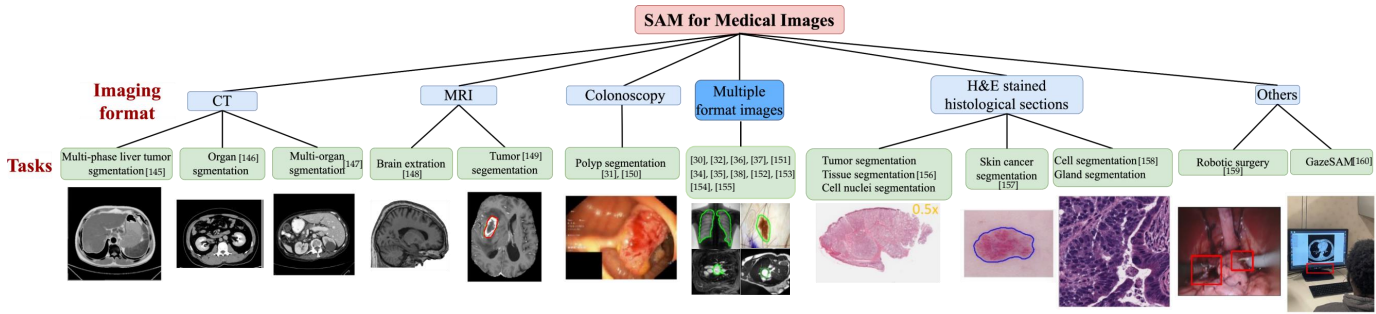


\subsubsection{Medical Imaging}
The goal of medical image segmentation is to reveal the anatomical or pathological structure of the corresponding tissues, which can assist computer-aided diagnosis and intelligent clinical surgery \cite{liu2022diagnosis,huang2022ultrasound}. Due to the rapid development of
computation power and medical data resources, deep learning-based medical image
segmentation has achieved massive progress in accuracy and speed against traditional counterparts \cite{chen2021uscl,gao2021multi}. With the emerging Visual Transformer (ViT), ViT-based medical image methods~\cite{chen2021transunet, cao2023swin, azad2022dae, azad2022transdeeplab} has achieved surpassing performance in the medical image segmentation. However, such networks are towards a specific task, which lacks the generalization ability on other tasks. Recently, SAM has been proposed to make it possible to solve multiple kinds of segmentation tasks in a unified framework. In this context, researchers have paid attention to customizing SAM for medical image segmentation and concluded some useful strategies to improve its
performance. This work~\cite{zhang2023segment} briefly summarizes recent efforts to extend the success of SAM to medical image segmentation tasks, while we provide a more comprehensive summary with deeper insight in this section.

According to the imaging format of medical images, the usage of SAM in medical image segmentation can be categorized into six series: Computerized Tomography (CT) images, Magnetic Resonance Imaging (MRI) images, colonoscopy images, H\&E stained histological sections images, multiple format images, and others. 

\textbf{For CT images.} A CT scan combines a series of X-ray images taken from different angles around your body and uses computer processing to create cross-sectional images (slices) of the bones, blood vessels, and soft tissues inside your body. This paper~\cite{hu2023sam} presents a preliminary examination of SAM as an annotation tool for medical image analysis, specifically for the segmentation of multi-phase liver tumors (MPLiTS). Their investigation focused on the prompts used, data resolution, and phases. The experimental findings demonstrate the effectiveness of SAM in this context, while also highlighting areas where MPLiTS could be improved. To provide comprehensive guidance to the MPLiTS community, the authors plan to conduct further investigations that encompass a wider range of aspects. The purpose of this paper~\cite{roy2023sam} is to perform an initial assessment of SAM's out-of-the-box zero-shot capabilities for medical image segmentation by evaluating its performance on an abdominal CT organ segmentation task using point or bounding box-based prompting. The findings indicate that SAM can generalize effectively to CT data, which could potentially accelerate the development of semi-automatic segmentation tools for clinicians. SAMed~\cite{zhang2023customized} is the proposed solution for medical image segmentation, which differs from previous methods in that it leverages the SAM, a large-scale image segmentation model. This approach involves customizing the SAM model for medical image segmentation by applying the low-rank-based (LoRA) finetuning strategy to the SAM image encoder. Unlike SAM, SAMed can perform better for semantic segmentation tasks on medical images. The trained SAMed model achieves comparable performance to SOTA methods. Additionally, since SAMed only updates a small fraction of the SAM parameters, its deployment and storage costs are minimal in practical usage.

\textbf{For MRI images.} MRI is a non-invasive diagnostic imaging technique that uses a powerful magnetic field, radio waves, and a computer to produce detailed images of internal structures in the body. MRI is commonly used to visualize the brain, spine, joints, and other soft tissues. This study~\cite{mohapatra2023brain} compares SAM with FSL's Brain Extraction Tool (BET), a widely used and current gold standard brain extraction technique, on a variety of brain scans with varying image qualities, MR sequences, and brain lesions affecting different brain regions. The findings show that SAM outperforms BET based on average Dice coefficient, IoU, and accuracy metrics, particularly in cases where image quality is compromised by signal inhomogeneities, non-isotropic voxel resolutions, or the presence of brain lesions near or involving the outer regions of the brain and meninges. Furthermore, SAM has superior segmentation properties, enabling a fine-grain separation of different tissue compartments and brain structures. These results suggest that SAM has the potential to become a more accurate, robust, and versatile tool for a broad range of brain extraction and segmentation applications. The paper~\cite{putz2023segment} demonstrates that SAM can achieve high segmentation accuracy for brain tumor MRI datasets in a point-to-mask setting, and effectively generalizes to brain tumor MRI datasets and achieves segmentation accuracy similar to what was observed in the 2D photographs where it was previously evaluated. Furthermore, the authors identify the challenges encountered when using SAM for tumor segmentation in MRI datasets and propose strategies to address them, which can also be applied in the context of clinical implementation.

\textbf{For colonoscopy images.} A colonoscopy is a test to check inside the bowels. This report~\cite{zhou2023can} evaluates the performance of SAM in segmenting polyps under unprompted settings. Polyp-SAM~\cite{li2023polyp} is a finetuned SAM model designed for polyp segmentation. The authors evaluate its performance against various SOTA polyp segmentation models and compare the performance of two transfer learning strategies: one involving finetuning of the encoders and the other without. Experimental results on five public datasets illustrate SOTA performance on two datasets and impressive performance on the remaining three.

\textbf{For H\&E stained histological sections images.} H\&E stained histological sections refer to tissue samples that have been processed and stained with hematoxylin and eosin (H\&E) dyes for microscopic examination. This staining technique is commonly used in histology and pathology to highlight different structures and cellular components within the tissue sample. This test can help find what's causing your bowel symptoms. This work~\cite{deng2023segment} assesses the zero-shot segmentation performance of the SAM model on representative segmentation tasks in whole slide imaging, including tumor segmentation, non-tumor tissue segmentation, and cell nuclei segmentation. The core results indicate that the zero-shot SAM model achieves remarkable segmentation performance for large connected objects. The authors also identify several limitations for digital pathology, including image resolution, multiple scales, prompt selection, and model fine-tuning. To address these limitations, few-shot fine-tuning with images from downstream pathological segmentation tasks may help the model achieve better performance in dense object segmentation in the future. This paper~\cite{zhang2023input} demonstrates that its generated masks, features, and stability scores can be leveraged to build and train more advanced medical image segmentation models. Specifically, it showcases how SAM can be used to enhance image inputs for a commonly used medical image segmentation model, such as U-Net. The proposed method is tested on two datasets, and the experiments demonstrate its effectiveness. SkinSAM~\cite{hu2023skinsam} presents a fine-tuned based model on SAM for skin cancer segmentation, which demonstrates remarkable segmentation performance. Experimental results also illustrate that larger models, such as ViT\_L and ViT\_H, can perform better than the smaller ViT\_b, the fine-tuned model (ViT\_b\_finetuned) showed the greatest improvement. 

\textbf{For multiple format images.} Works in this part evaluate SAM or propose SAM-based segmentation methods for more than one segmentation task with multiple kinds of medical images. 

\begin{itemize}
\item \textbf{Evaluation of SAM’s ability to segment medical images.} This paper~\cite{ma2023segment} is the first work at extending the success of SAM to medical images. It compiles an extensive medical image dataset comprising 11 distinct modalities and containing more than 200,000 masks. This work~\cite {mazurowski2023segment} is an extensive evaluation of SAM’s ability to segment medical images on a collection of 11 medical imaging datasets from various modalities and anatomies.  This paper~\cite{he2023accuracy} explores the accuracy of SAM on 12 public medical image segmentation datasets
which cover various organs (brain, breast, chest, lung, skin, liver, bowel, pancreas, and prostate), image modalities (2D X-ray, histology,
endoscopy, 3D MRI, and CT), and health conditions (normal, lesioned). This work~\cite{shi2023generalist} evaluates SAM on medical images and presents both quantitative and qualitative results of zero-shot segmentation on nine benchmarks for medical image segmentation. These benchmarks encompass a variety of imaging modalities, including optical coherence tomography (OCT), magnetic resonance imaging (MRI), and computed tomography (CT), and different applications such as dermatology, ophthalmology, and radiology. This paper~\cite{cheng2023sam} evaluates SAM's zero-shot generalization on medical images by collecting more than 12 public medical image datasets that cover various organs and modalities. This paper~\cite{mattjie2023exploring} evaluates the zero-shot performance of SAM 2D in medical imaging by testing it on six datasets from four different imaging modalities, including X-ray, ultrasound, dermatoscopy, and colonoscopy. The findings indicate that SAM 2D's zero-shot performance is either comparable or superior to the existing SOTA models. Huang \etal~\cite{huang2023segment} collect and sort 52 open-source datasets and build a large medical segmentation dataset with 16 modalities, 68 objects, and 553K slices. A comprehensive analysis of different SAM testing strategies on the so-called COSMOS 553K dataset is conducted.

\item \textbf{SAM-based segmentation methods for medical images.} This paper~\cite{ma2023segment} proposes a straightforward fine-tuning approach to tailor the SAM model for general medical image segmentation. Rigorous experimentation on 21 3D segmentation tasks and 9 2D segmentation tasks illustrates that MedSAM surpasses the default SAM model. Segment Any Medical Model (SAMM)~\cite{liu2023samm} is an extension of SAM on 3D Slicer, a widely-used open-source image processing and visualization software that has been extensively used in the medical imaging community. Medical SAM Adapter (MSA)~\cite{wu2023medical} first integrates the medical-specific domain knowledge to the segmentation model and shows superior
performance on 19 medical image segmentation tasks with various image modalities, including CT, MRI, ultrasound image, fundus image, and dermoscopic images. This paper~\cite{qiu2023learnable} introduces Learnable Ophthalmology SAM, a novel approach to multiple target segmentation in ophthalmology multimodal images. The proposed method incorporates a learnable prompt layer that extracts medical prior knowledge from each transformer layer. Using a one-shot mechanism during training, this work only trains the prompt layer and task head. Experiments are conducted to demonstrate the effectiveness of the proposed method on four medical segmentation tasks (\ie blood vessel segmentation, lesion segmentation, and retinal layer segmentation) across nine publicly available datasets. SAM-Adapter~\cite{chen2023sam} leverages domain-specific information or visual prompts to enhance the segmentation network through the use of simple yet effective adapters. By combining task-specific knowledge with general knowledge learned by the large model, SAM-Adapter can notably improve the performance of SAM in challenging tasks, as demonstrated by extensive experiments. PromptUNet~\cite{junde2023PromptUNet} is proposed by extending the existing prompt types in SAM to include novel Supportive Prompts and En-face Prompts. The authors evaluate its capabilities on 19 medical image segmentation tasks using a variety of image modalities and PromptUNet exceeds a wide range of state-of-the-art (SOTA) medical image segmentation methods.
\end{itemize}

\textbf{For other medical images.} This paper~\cite{wang2023sam} evaluates SAM's performance using two established robotic instrument segmentation datasets from the MICCAI EndoVis 2017 and 2018 challenges. The extensive evaluation indicates that SAM displays remarkable zero-shot generalization ability with bounding box prompts. Moreover, the qualitative figures demonstrate that the model either fails to predict certain parts of the instrument mask (such as the jaws and wrist) or misclassifies parts of the instrument in scenarios where instruments overlap within the same bounding box or with point-based prompts. However, in some complex surgical scenarios involving blood, reflection, blur, and shade, SAM is unable to identify instruments. Additionally, SAM's performance is insufficiently robust to withstand various forms of data corruption. GazeSAM~\cite{wang2023gazesam} is the first work to leverage the power of eye-tracking technology
and SAM to enhance the efficiency of daily clinical practice. It enables radiologists to collect segmentation masks by simply looking at the region of interest during image diagnosis. This system tracks the radiologists' gaze and uses the resulting data as input prompts for SAM, which then generates the segmentation mask in real-time automatically. 
\subsubsection{Video}
In the field of computer vision, Video Object Tracking (VOT) \cite{kristan2021ninth, zhang2022webuav} and video segmentation are recognized as crucial and indispensable tasks. It involves locating a particular object in a video frame and subsequently tracking it throughout the rest of the video. As such, it has various practical applications, such as surveillance and robotics.

\begin{figure}[htbp]
    \centering
    \includegraphics[width=0.495\textwidth]{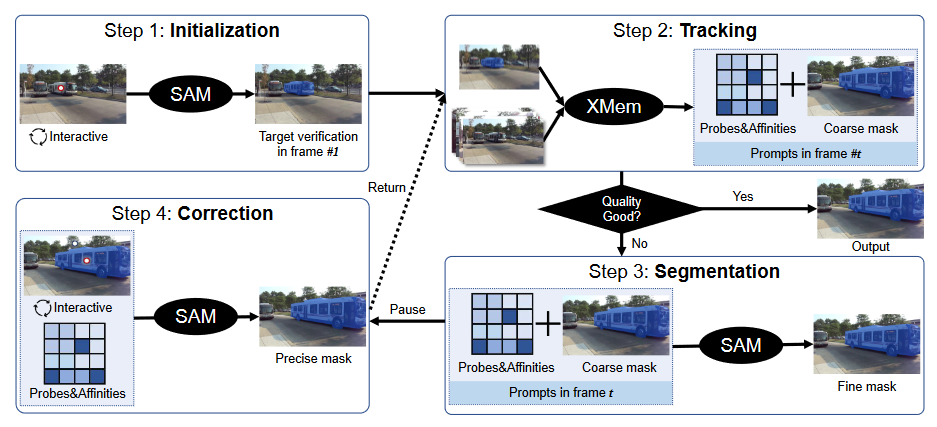}
    \caption{Pipeline of TAM. The figure is borrowed from the original paper~\cite{yang2023track}.}
    \label{fig:tracking_anything.png}
\end{figure}

SAM \cite{sam} has made outstanding contributions in the field of VOT. Track Anything Model (TAM) \cite{yang2023track} was introduced, which achieves outstanding interactive tracking and segmentation in videos with remarkable performance. This research paper suggests the utilization of a highly-effective toolkit named Track-Anything for high-performance object tracking and segmentation in videos. Unlike current existing methods, it adopts an interactive approach for initialization and incorporates the SAM and XMem \cite{cheng2022xmem}. The pipeline of  TAM is illustrated in Fig.~\ref{fig:tracking_anything.png}. SAM is a large segmentation model with powerful segmentation abilities, while XMem is an advanced semi-supervised VOS model. The proposed method exhibits exceptional performance and user-friendliness in complex situations, with potential applications in many fields. Additionally, this paper scrutinizes failed instances and recommends optimal solutions. The proposed method has the ability to tackle challenging situations in video object perception with efficiency.

\begin{figure}[htbp]
    \centering
    \includegraphics[width=0.495\textwidth]{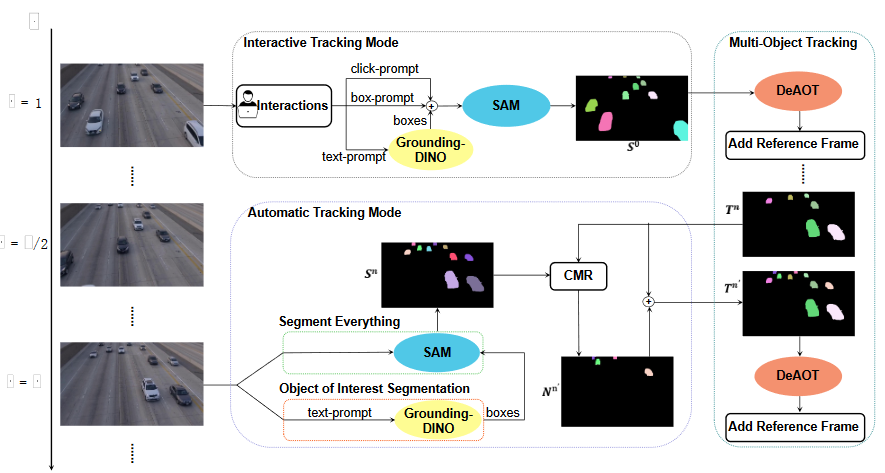}
    \caption{Pipeline of SAM-Track. The figure is borrowed from the original paper~\cite{Cheng2023SegmentAT}.}
    \label{fig:SenTA.png}
\end{figure}

Furthermore, another tracking model, namely SAM-Track~\cite{Cheng2023SegmentAT} has been proposed. SAM-Track is a video segmentation framework that enables object tracking and segmentation through both interactive and automatic methods. The pipeline of  SAM-Track is illustrated in Fig.~\ref{fig:SenTA.png}. SAM-Track combines Grounding-DINO\cite{liu2023grounding}, DeAOT\cite{yang2022decoupling}, and SAM to achieve these capabilities. By employing both interaction and automation, SAM-Track becomes a highly versatile tool with potential beyond just object tracking and segmentation. Its combination of features has allowed it to achieve successful results, demonstrating its impact.

\begin{figure}[htbp]
    \centering
    \includegraphics[width=1.0\linewidth]{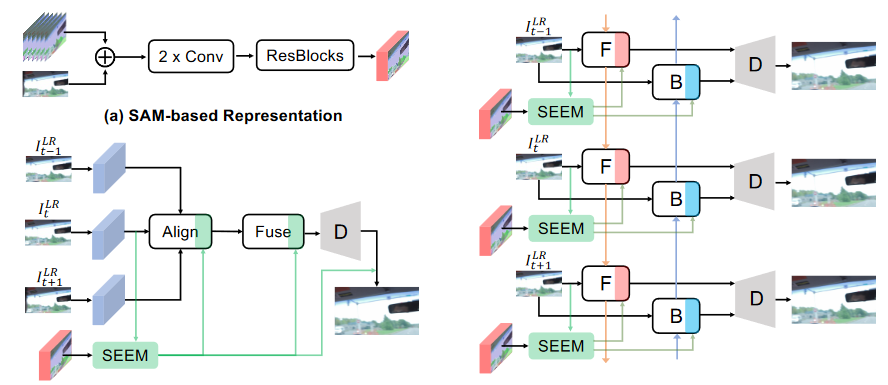}
    \caption{Overview of the proposed framework from the original paper~\cite{Lu2023CanSB}. (a) Illustrate how to obtain SAM-based representation. (b) Apply the proposed SAM-guided refinement module (SEEM) to the sliding-window-based method. (c) Apply SEEM to the bidirectional recurrent structure-based method.}
    \label{fig:cansamboost.png}
\end{figure}

In addition to its applications in tracking tasks, SAM has also shown promise in enhancing the resolution of video. Video super-resolution (VSR) \cite{fuoli2019efficient}\cite{haris2019recurrent}\cite{huang2015bidirectional}\cite{huang2017video}\cite{huilightweight} presents a primary challenge of handling large motions in input frames, despite its potential in accurately aggregating information from multiple frames. This challenge requires accurate displacement estimation of input frames in complex scenes, which is a challenge both in terms of real-time processing and computational complexity. \cite{Lu2023CanSB} proposes a method to enhance the quality of video super-resolution (VSR) by utilizing a SAM to build a more robust and semantically-aware prior. Moreover, the authors have devised a lightweight module called SAM-guided refinement module (SEEM) to boost the performance of existing methods. The framework of  SEEM is illustrated in Fig.~\ref{fig:cansamboost.png}. The SEEM module is easy to integrate into existing methods. Experimental results show that SEEM can provide superior performance with both comprehensive fine-tuning and fine parameter adjustment.

\subsubsection{Data Annotations}
Data annotation in AI involves the process of labeling data for machine learning algorithms to help them learn to recognize specific patterns, objects, or features. Accurate data annotation is crucial to the development of effective machine learning models that can successfully perform tasks like object detection, classification, and natural language processing. Due to the high cost associated with annotating images and videos in certain fields, many datasets have not been labeled effectively, particularly at the pixel level. However, the emergence of SAM\cite{sam} will facilitate the effective labeling of such datasets.

SAMText\cite{he2023scalable} is a scalable pipeline for mask annotation of scene text in videos. The pipeline utilizes the SAM for generating mask annotations in a large-scale dataset, SAMText-9M, which contains over 2,400 video clips and more than 9 million mask annotations. The authors argue that annotating scene text at a finer level can lead to remarkable improvements in detection and recognition performance, even for curved text. Furthermore, the paper identifies several potential research directions, such as examining the effect of mask annotations, enhancing data and model scalability, and generating character-level mask annotations.

\cite{wang2023scaling} presents a method to construct a large-scale remote sensing image segmentation dataset named SAMRS by utilizing existing remote sensing object detection datasets and the data-centric machine learning model SAM. SAMRS includes object category, location, and instance information, which can be utilized for semantic segmentation, instance segmentation, and object detection research. The authors conducted a comprehensive analysis of the dataset and discussed SAM's performance in remote sensing image segmentation, demonstrating its potential to improve annotation efficiency. SAMRS surpasses previously existing high-resolution remote sensing image segmentation datasets in size. SAMRS is a valuable resource when conducting large-scale model pre-training research in the field of remote sensing image segmentation.

Generating high-quality pseudo-labels is a crucial step towards achieving accurate and reliable results in various computer vision applications. These pseudo-labels can be used to train and test various computer vision models, enabling researchers and practitioners to achieve state-of-the-art results in object recognition, semantic segmentation, and other related tasks. SAM has made it incredibly easy, fast, and highly effective to generate high-quality pseudo-labels\cite{Chunming2023WSCOS}\cite{chen2023learning}\cite{jiang2023segment}, thus facilitating new breakthroughs in the field of computer vision research and applications.

He \etal~\cite{Chunming2023WSCOS} propose a new method for weakly-supervised multi-object semantic segmentation. The WS-SAM framework utilizes the recently proposed vision foundation model, SAM, to generate segmentation masks, and proposes several techniques, including multi-augmentation result fusion, pixel-level uncertainty weighting, and image-level uncertainty filtration, to obtain reliable pseudo labels for training a segmentation model. Chen \etal~\cite{chen2023learning} introduce a framework for pretraining thermal infrared image segmentation models using pseudo-labels generated with the SAM model. The framework can effectively improve the accuracy of segmentation results for specific categories, exceeding the SOTA ImageNet pretrained model. The authors generated a thermal infrared segmentation dataset containing over 100,000 images for pretraining. Jiang \etal~\cite{jiang2023segment} present an approach to weakly supervised semantic segmentation that utilizes cheap annotations such as image labels, points, and scribbles as prompts for SAM. This model outputs object masks with precise boundaries, which are used to generate pseudo labels for training segmentation networks. The experiments on the PASCAL VOC\cite{everingham2015pascal} show that SAM can serve as an effective pseudo-label generator.

\subsection{Beyond Vision}
\subsubsection{3D Reconstruction}
\begin{figure}[htbp]
    \centering
    \includegraphics[width=0.98\linewidth]{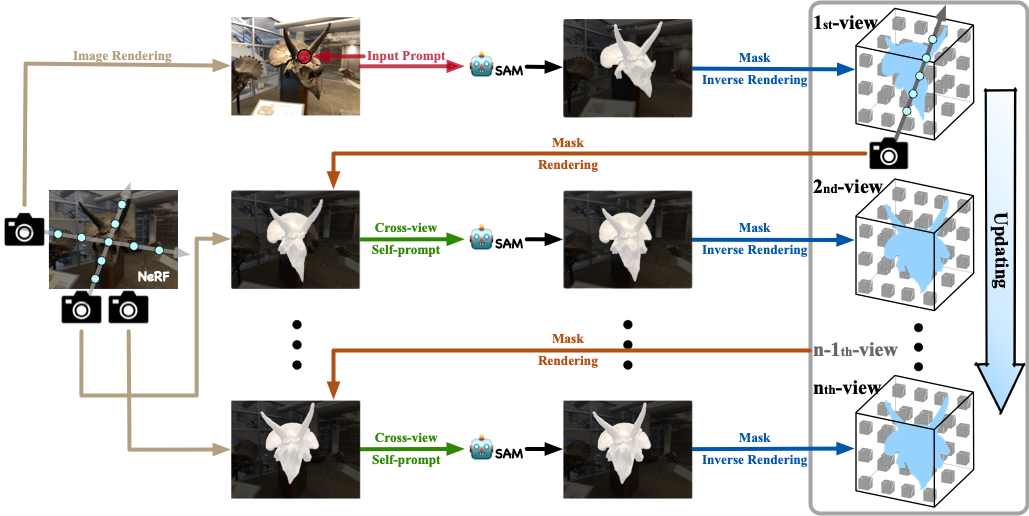}
    \caption{Overall pipeline of SA3D~\cite{3DNERFSA}. Initially, the user provides prompts for the target object, and SAM generates a segmentation mask for it. This 2D mask is then projected onto a 3D mask grid using mask inverse rendering, taking into account the density distribution. By rendering incomplete 2D masks from novel views, SA3D extracts reliable prompts and queries SAM to obtain accurate segmentation masks. This iterative process, called cross-view self-prompting, alternates between mask inverse rendering and obtaining segmentation masks, ultimately resulting in the final 3D segmentation results.}
    \label{fig:SA3D_pipeline}
\end{figure}
In addition to achieving fine-grained 3D segmentation, SA3D\cite{3DNERFSA} can also be used for 3D reconstruction. With the 3D mask grid obtained from the previous section, we can determine the occupied space of objects in 3D and reconstruct them in various ways.

However, due to the high memory requirements and computational complexity of NeRF-based methods, they are currently limited to relatively small scenes and cannot handle large-scale outdoor scenes. To tackle this challenge, several studies~\cite{PiFu, NeRF++} have proposed to use additional input modalities such as depth maps and surface normals to improve the efficiency and accuracy of NeRF-based 3D reconstruction.

SAM\cite{sam} is a SOTA method for 2D image segmentation that can segment anything with user-specified prompts. SAM can be used in various applications such as object detection, image retrieval, and image synthesis. However, SAM is currently limited to 2D image data and cannot be directly applied to 3D scene understanding.

The proposed SA3D framework, as shown in Fig.~\ref{fig:SA3D_pipeline} extends the segmentation ability of SAM to 3D scenes by leveraging NeRFs. SA3D can segment any object in a 3D scene with one-shot manual prompting in a single rendered view. SA3D utilizes mask inverse rendering and cross-view self-prompting techniques to project 2D masks onto 3D mask grids and generate new prompts for different views, respectively. Compared to previous approaches based on NeRFs, SA3D can easily adapt to any pre-trained NeRF without any changes and re-training.

\begin{figure}[htbp]
    \centering
    \includegraphics[width=0.98\linewidth]{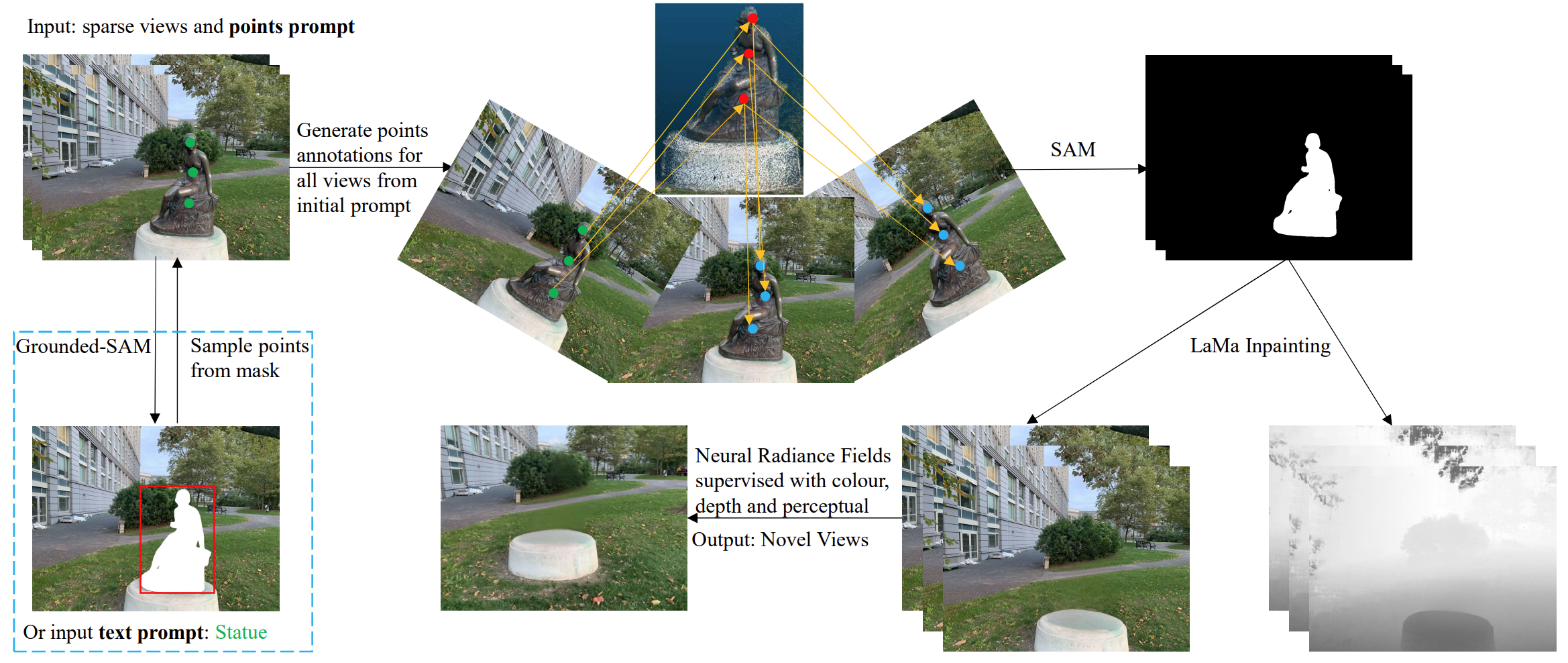}
    \caption{Framework of OR-NeRF\cite{yin2023ornerf}.}
    \label{fig:OR-NeRF}
\end{figure}
On the contrary of reconstructing objects, this paper\cite{yin2023ornerf} presents a new object-removing pipeline called OR-NeRF that removes objects from 3D scenes using either point or text prompts on a single view. The method achieves better performance in less time than previous works by using a points projection strategy to rapidly spread user annotations to all views. The algorithm shown in Fig.~\ref{fig:OR-NeRF} leverages the recent 2D segmentation model SAM\cite{sam} to predict masks with improved precision and efficiency and obtains color and depth priors through 2D inpainting methods. Finally, the algorithm employs depth supervision and perceptual loss for scene reconstruction to maintain consistency in appearance after object removal.

Although the above methods have demonstrated promising performance in 3D segmentation, there is still much room for improvement. Here, we discuss several potential future directions for 3D reconstruction:

\begin{itemize}
\item \textbf{Improving the efficiency of NeRF-based methods:} NeRF-based methods are computationally expensive and memory-intensive, limiting their scalability to large-scale scenes. Future research can focus on improving the efficiency and scalability of NeRF-based methods to enable real-time 3D scene understanding.
\item \textbf{Incorporating additional input modalities:} In addition to multi-view images, incorporating additional input modalities such as depth maps and surface normals can improve the accuracy and efficiency of NeRF-based 3D reconstruction and segmentation.
\item \textbf{Integrating with other 3D perception tasks:} 3D segmentation is an important task in 3D scene understanding. Integrating SA3D with other 3D perception tasks, such as 3D object detection, 3D scene reconstruction, and 3D pose estimation, can lead to more comprehensive 3D scene understanding.
\item \textbf{Expanding the scope of applications:} SA3D can be applied to various applications such as robotics, augmented reality, and virtual reality. Future research can explore the potential of SA3D in these fields and develop novel applications.
\end{itemize}

\subsubsection{Non-Euclidean Domain}

\begin{figure}[htbp]
  \centering
  \includegraphics[width=1.0\linewidth]{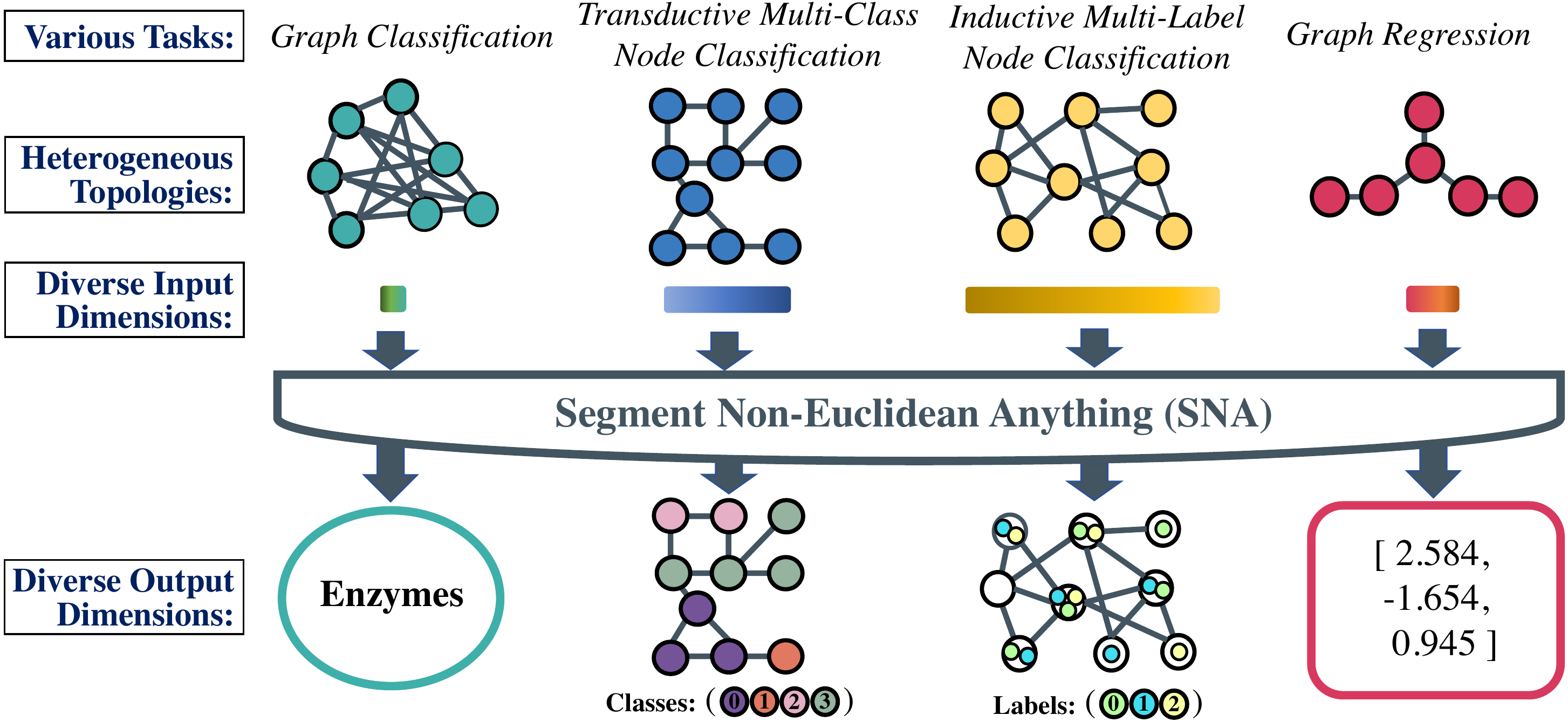}
  \caption{Illustrations of SNA from the
original paper~\cite{Non-euclidean_segment}.}
  \label{fig:non_euclidean_SA}
\end{figure}

In the context of graph neural networks, the Non-Euclidean domain refers to graphs that are irregular and do not have a predefined structure like grids or lattices. These graphs can represent a wide range of data, including social networks, citation networks, e-commerce product graphs, and molecule graphs. Due to the complexity and heterogeneity of these graphs, developing a foundation model for universal graph analysis has become a challenging task.

In recent years, several research works have been conducted to address the challenges of graph analysis in the Non-Euclidean domain. For example, the work presented in \cite{kipf2016semi} proposes the Graph Convolutional Network (GCN), which is based on spectral graph convolutions and can be applied to a wide range of graph-related tasks. The GraphSAGE method proposed by Hamilton \etal~\cite{hamilton2017inductive} enables the scalable processing of large graphs by sampling neighbors instead of using all of them. The Graph Attention Network (GAT) introduced by Velivckovic \etal~\cite{velivckovic2017graph} utilizes the attention mechanism to learn the weights for each neighbor automatically.

Despite the progress made in Non-Euclidean graph analysis, there is still a need for more versatile and adaptable foundation models. In particular, the recent work presented in~\cite{sam} introduces the SAM, which is a prompt-based framework for general image analysis. SAM enables users to input natural language prompts to perform a range of image-related tasks, including object detection, segmentation, and classification. Inspired by the success of SAM, the Segment Anything in Non-Euclidean (SNA)~\cite{Non-euclidean_segment} paradigm aims to develop a foundation model for universal graph analysis that is flexible, adaptable, and capable of handling diverse graph samples and tasks.

To address the challenges of handling diverse feature dimensions for varying tasks, the SNA approach shown in Fig.~\ref{fig:non_euclidean_SA} introduces a specialized slimmable graph convolutional layer. This layer enables the dynamic activation or deactivation of its channels based on the input feature dimensions. Furthermore, the approach proposed in this paper employs a meta-learning strategy that learns to select the optimal neurons based on the downstream tasks rather than relying on manual selection. This approach enables the handling of a broader range of unseen tasks and represents a step towards developing more versatile and adaptable foundation models for universal graph analysis.

In summary, the SNA paradigm aims to establish a foundation model for universal graph analysis in the Non-Euclidean domain. The SNA approach leverages the prompt-based framework introduced in the SAM and introduces specialized slimmable graph convolutional layers and a meta-learning strategy to handle the challenges of diverse graph samples and tasks. The proposed approach has the potential to inspire future research in developing more versatile and adaptable foundation models for graph analysis in the Non-Euclidean domain.

\subsubsection{Robotics}
In recent years, the use of foundation models has advanced various applications such as image segmentation and natural language processing. In this paper\cite{huang2023instruct2act}, the authors present Instruct2Act, a framework that utilizes Large Language Models (LLMs) to map multimodal instructions to sequential actions for robotic manipulation tasks. The framework employs the LLM model to generate Python programs that form a comprehensive perception, planning, and action loop for robotic tasks.

\begin{figure}[htbp]
\centering
\includegraphics[width=0.5\textwidth]{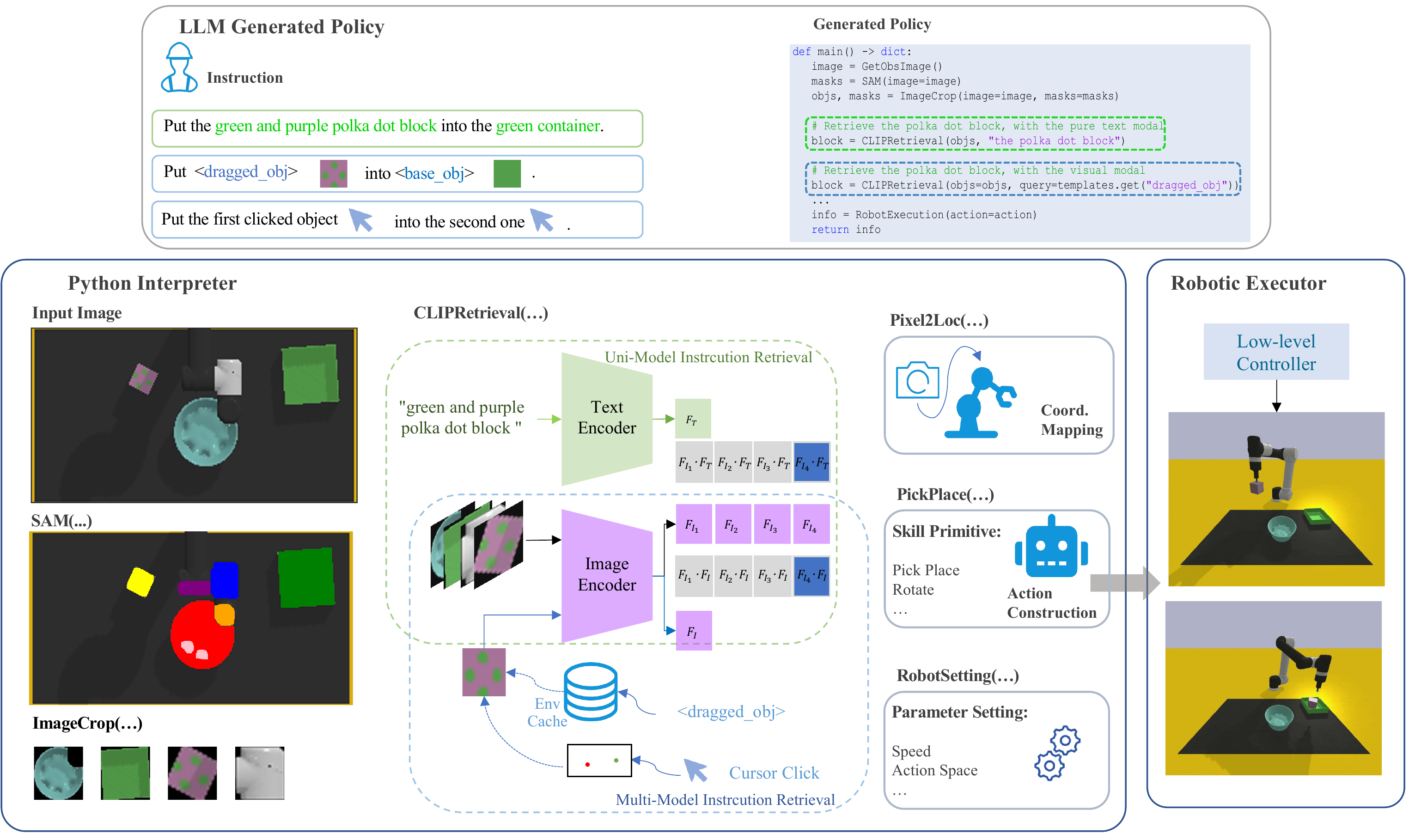}
\caption{Paradigm of the Instruct2Act\cite{huang2023instruct2act} framework. Best viewed by zooming in.}
\vspace{-0.3cm}
\label{fig:Instruct2Actframewordframework}
\end{figure}

Fig.~\ref{fig:Instruct2Actframewordframework} shows the overall pipeline of Instruct2Act\cite{huang2023instruct2act}. In the perception section, pre-defined APIs are used to access multiple foundation models. The SAM\cite{sam} accurately locates candidate objects, and CLIP\cite{radford2021learning} classifies them. The framework leverages the expertise of foundation models and robotic abilities to convert complex high-level instructions into precise policy codes. In this way, Instruct2Act\cite{huang2023instruct2act} is adjustable and flexible in accommodating various instruction modalities and input types, catering to specific task demands.

The practicality and efficiency of Instruct2Act\cite{huang2023instruct2act} were validated by assessing it on robotic tasks in different scenarios within tabletop manipulation domains. Moreover, the zero-shot method outperformed many state-of-the-art learning-based policies in several tasks. The code for Instruct2Act\cite{huang2023instruct2act} is available at https://github.com/OpenGVLab/Instruct2Act, serving as a robust benchmark for high-level robotic instruction tasks with assorted modality inputs. This framework provides a promising approach for enabling robots to perform complex tasks by leveraging the power of foundation models and large language models.

In conclusion, Instruct2Act\cite{huang2023instruct2act} sets foot into the robotics area with the latest powerful tools for the first time, this may be a foot print of humanity stepping into the AGI. We will see more research in this field, paving the way for a more productive society.

\subsubsection{Video Text Spotting}
Video text spotting is a challenging task that involves localizing and recognizing text instances in video frames or sequences. Traditional methods for video text spotting have relied on the detection of bounding boxes and the subsequent recognition of text instances within those boxes. However, these methods are limited in their ability to accurately localize text instances, particularly those with irregular shapes or orientations.

\begin{figure}[htbp]
  \centering
    \includegraphics[width=1\linewidth]{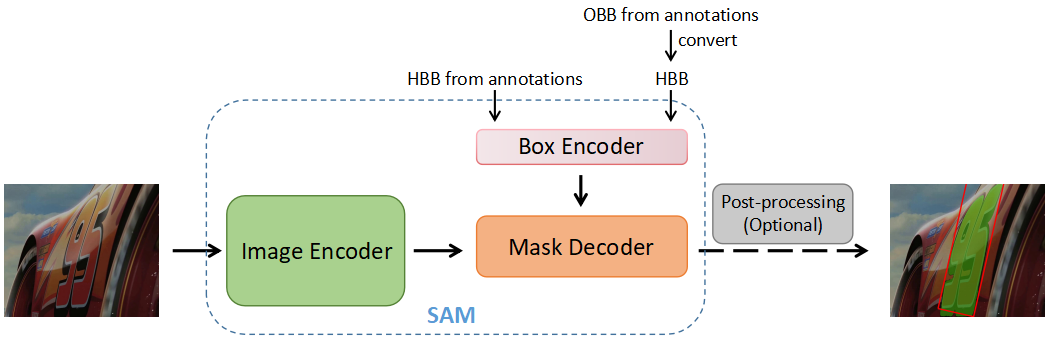}
    \caption{Overview of SAMText~\cite{he2023scalable} pipeline, which extends the SAM approach to generate mask annotations for scene text images or video frames in a scalable manner. The pipeline takes an input bounding box, which can either come from existing annotations or be derived from a scene text detection model. }
\label{fig:samtext_stat_dataset-pipeline}
\end{figure}

Recent advances in computer vision, such as the segmentation-based approach proposed by the SAM~\cite{sam}, have shown promise in addressing these limitations. The SAM approach utilizes a deep neural network to generate pixel-level segmentation masks for text instances, resulting in more accurate and fine-grained annotations.

In this context, the SAMText~\cite{he2023scalable} pipeline shown in Fig.~\ref{fig:samtext_stat_dataset-pipeline} offers a scalable and efficient solution for generating mask annotations for video text spotting tasks. By applying the SAM model to bounding box annotations, SAMText is able to generate mask annotations for large-scale video text datasets, as demonstrated by the SAMText-9M dataset.

While SAMText is a novel approach for generating mask annotations for video text spotting tasks, it builds upon the foundation laid by the SAM model. Specifically, the SAM model's ability to generate high-quality pixel-level masks for objects in images has been adapted for the specific task of generating masks for text instances in video frames.

Given an input scene text image or video frame, SAMText begins by extracting the bounding box coordinates from existing annotations or derived from a scene text detection model. If the boxes are oriented, SAMText calculates their minimum bounding rectangle to obtain the horizontal bounding boxes (HBB), which is then used as the input prompt for the SAM model to obtain mask labels. The SAM model is a segmentation model that is pre-trained on natural images and fine-tuned on the COCO-Text dataset to generate mask annotations for text instances.

After obtaining the mask for each text instance, post-processing may be necessary to ensure its connectivity. In particular, if a mask comprises several segments, it may be desirable to derive a minimum enclosing mask as an optional step in order to achieve a more cohesive representation. Furthermore, optical flow estimation can also be utilized to enhance the accuracy of the generated masks and ensure temporal consistency.

The SAMText pipeline provides an exciting avenue for future research in video text spotting. By providing fine-grained mask annotations for large-scale datasets, SAMText enables the development and evaluation of more accurate and effective video text spotting models. Additionally, the SAMText approach may inspire the development of new segmentation-based approaches for other computer vision tasks.

Overall, the SAMText pipeline represents an important contribution to the field of video text spotting, offering a scalable and efficient solution for generating fine-grained mask annotations. The approach holds promise for advancing the accuracy and effectiveness of video text spotting models and may inspire further research in the field of computer vision.

\subsubsection{Vision and Language}

\begin{figure}[htbp]
    \centering
    \includegraphics[width=1\linewidth]{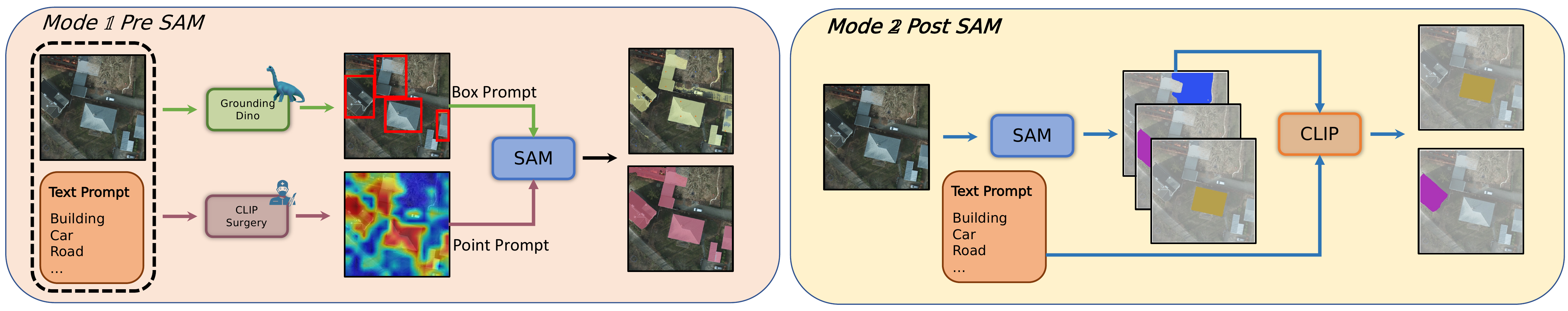}
    \caption{Overall structure of the Text2Seg~\cite{text2seg} pipeline, which consists of three methods for guiding the SAM model. The pipeline begins by using a text prompt as input for Grounding DINO, which generates bounding boxes. These bounding boxes are then passed to SAM, which produces a segmentation map. In the second step, a text prompt is given to CLIP Surgery, resulting in a heatmap. This heatmap is sampled to create point prompts for SAM, which generates segmentation masks. Finally, SAM is used to generate segmentation maps, and CLIP is employed to compare the semantic similarity between these maps and the text prompt. The figure is borrowed from the original paper~\cite{text2seg}.}
    \label{fig:text2seg_framework}
\end{figure}

SAM also benefits vision and language tasks, such as image caption and text-based segmentation. InternGPT\cite{liu2023interngpt} is a system that combines chatbots capable of planning and reasoning with non-verbal cues like pointing movements, allowing users to manipulate images and videos on a screen directly. The system uses SAM to segment objects.  In this study\cite{text2seg}, the authors propose a pipeline called Text2Seg, which leverages multiple visual foundation models to facilitate remote sensing image semantic segmentation tasks guided by text prompts. The authors focus on the remote sensing domain, where images are notably dissimilar from those in conventional scenarios, and traditional models often underperform when confronted with testing data collected under different scenarios.

The authors incorporate multiple visual foundation models into their pipeline, including the SAM, which was introduced by Meta AI Research as the first foundation model for the object segmentation task. SAM is capable of performing zero-shot segmentation using various visual prompts as guidance, making it particularly suitable for remote sensing imagery processing, where labeled datasets are often sparse and inherently heterogeneous.

However, adapting SAM for specific tasks remains a non-trivial task, as the model segments all objects into distinct masks, making it impractical to use directly for semantic segmentation tasks. To address this issue, the authors propose leveraging other foundation models that focus on different aspects to generate visual prompts as guidance for the SAM model. These models can generate points or bounding boxes that help narrow down the areas where SAM needs to make predictions or assist in filtering SAM's predictions based on specific text prompts.

The authors evaluated their pipeline (shown in Fig.~\ref{fig:text2seg_framework}) on four commonly used remote sensing datasets and achieved promising results. They showed that although SAM is effective in segmenting instances within the provided frame, generating segmentation masks for a specific category remains challenging. In contrast, other visual foundation models, like Grounding DINO\cite{dino} and CLIP\cite{radford2021learning}, exhibit a superior capacity to understand the semantic features of images, thus generating coarse-grained visual prompts. Integrating these models into a unified pipeline allows us to exploit their combined strengths.

Overall, this work provides insights into maximizing the applicability of visual foundation models in specific contexts with minimal model tuning. The authors' proposed pipeline is not limited to any specific dataset and can be applied to various scenarios with minimal adjustments to the prompt tuning process. The authors hope that their work will encourage additional research on the application of visual foundation models for diverse downstream tasks and stimulate the development of increasingly powerful visual foundation models.

\begin{figure}[htbp]
    \centering
    \includegraphics[width=1.0\linewidth]{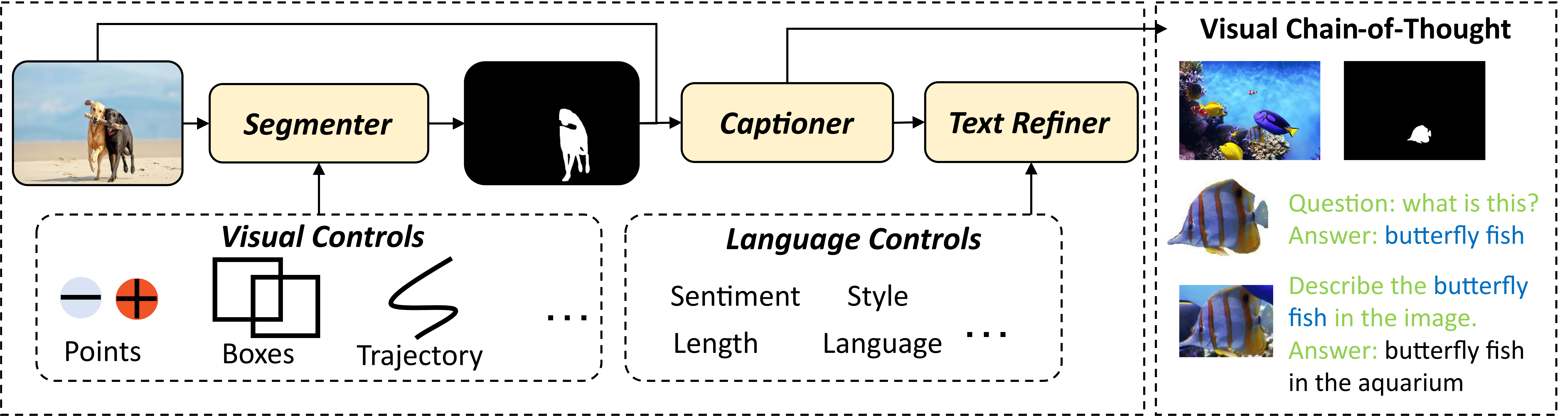}
    \caption{
Overview of the Caption Anything~\cite{wang2023caption} framework, which enhances image captioning by incorporating multimodal controls that align with human intention, allowing for a diverse range of visual focuses and language styles. The visual prompt is initially converted into a mask prompt using the \textit{segmenter}. Then, the \textit{captioner} generates a raw caption specifically for the region outlined by the mask. To ensure that the \textit{captioner} emphasizes the user's intended object, a straightforward visual chain-of-thought technique is employed for step-by-step inference. Finally, both the text prompt and the raw caption are input into the \textit{text refiner}, which generates a user-preferred caption that aligns with the desired genre.}
    \label{fig:captionanythingdemo}
\end{figure}

\textbf{Image captioning} is the task of generating natural language descriptions for given images. It is a fundamental problem in computer vision and natural language processing, with various applications in robotics, image retrieval, and content-based image retrieval \cite{wang2023memory}. In recent years, significant progress has been made in this field, especially with the development of deep learning techniques.

One recent approach for controllable image captioning is Caption Anything (CAT), proposed by Wang \etal~\cite{wang2023caption}. The framework, as shown in Fig.~\ref{fig:captionanythingdemo}, introduces multimodal controls to image captioning, rendering a variety of visual focuses and language styles aligned with human intention. The approach is formulated as a triplet solver, consisting of a \textit{segmenter}, a \textit{captioner}, and a \textit{text refiner}. The \textit{segmenter} takes the interactive visual controls and represents the user-interested regions via pixel-level masks, which are subsequently used by the \textit{captioner} to generate raw descriptions in relation to the specified region based on the original image and the provided mask. To facilitate the \textit{captioner} to focus on the user-interested object, the authors design a visual chain-of-thought technique with step-by-step inference. Finally, the \textit{text refiner} refines the raw descriptions by incorporating user-defined language controls, tailoring the language style according to user preferences.

The CAT approach leverages the SAM \cite{sam} to perform the segmentation task. SAM is a promptable segmentation model that achieves strong performance across various image domains. Specifically, SAM adapts interactive segmentation to achieve promptable ability, where a prompt, any interaction (\eg points, boxes) indicating what to segment in an image, is used to prompt SAM to return a valid segmentation mask. Once the user-specified segmentation mask is obtained, it is easy to generate the desired caption according to the original image and mask prompt.

The CAT approach is a training-free and adaptable solution for controllable image captioning tasks. It expands the range of supported control signals, enhances the model's flexibility and scalability, and offers strong user-interactive capabilities. The work highlights the importance of multimodal controls and prompts in controllable image captioning and provides insights into potential future research directions in this field.

In conclusion, vision and language tasks may not need training models anymore with the help of SAM and LLMs. We are about to see more powerful tools utilizing large pre-trained models or APIs to solve various kinds of tasks.

\subsubsection{Audio and Vision}
\begin{figure}[htbp]
    \centering
    \includegraphics[width=1.0\linewidth]{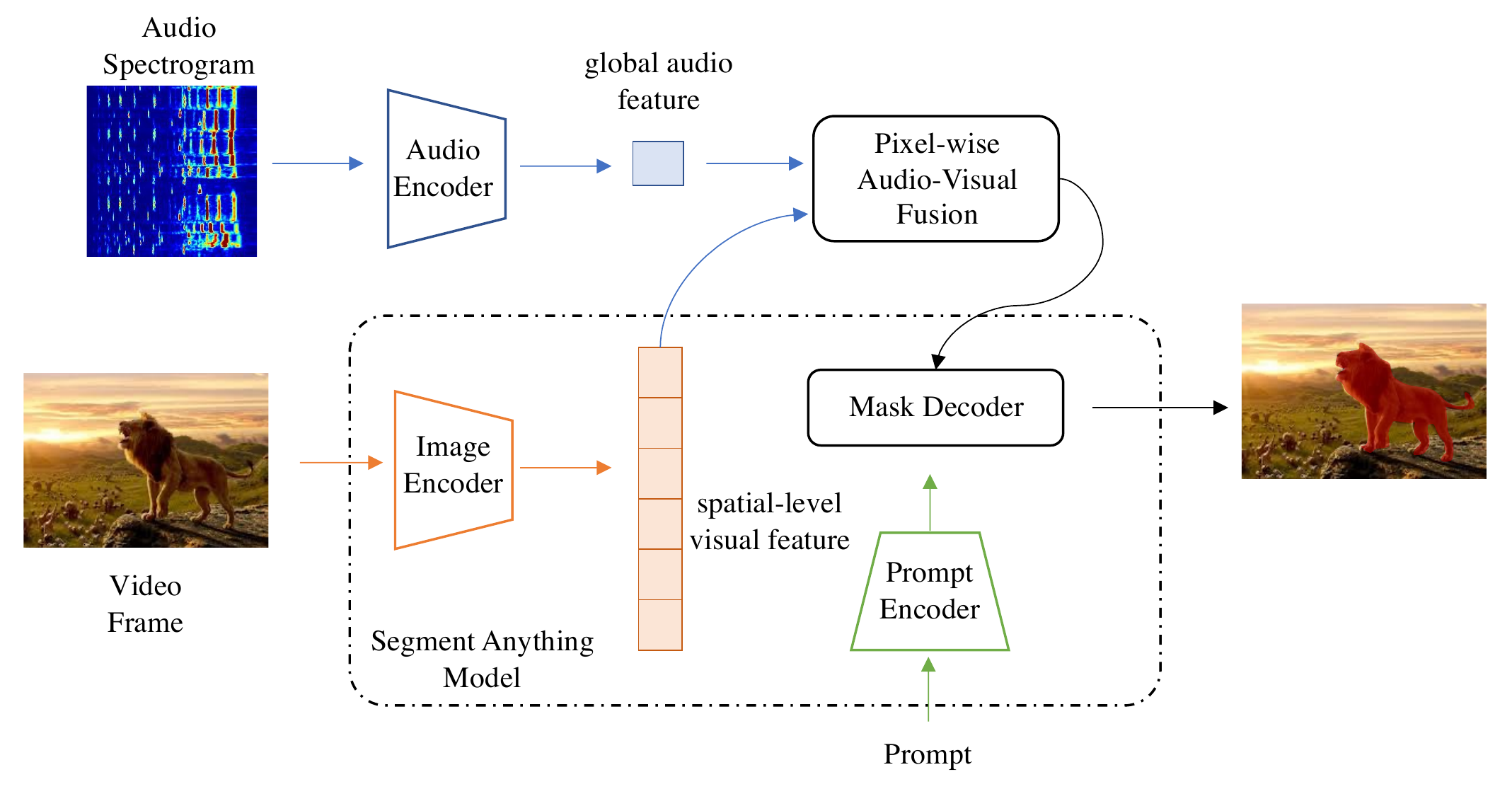}
    \vspace{-0.5em}
    \caption{Illustration of the proposed Segment Anything Model for Audio-Visual localization and segmentation (AV-SAM) from the original paper~\cite{mo2023av}.
The pixel-wise audio-visual fusion module combines audio features and visual features from pre-trained audio and image encoders. These cross-modal representations are aggregated, and along with prompt embeddings from the prompt encoder, they are fed into the mask decoder. The mask decoder utilizes this information to generate the final segmentation mask.
    }
    \label{fig: audio_visual_SA}
    \vspace{-1em}
\end{figure}

Audio and vision are two modalities that are closely related and can provide complementary information to solve many problems. In recent years, there has been an increasing interest in joint audio-visual learning \cite{liu2020re,liu2018visual}, which aims to learn the correlation between the two modalities and leverage the complementary information for better performance in various tasks.

One of the most popular applications of audio-visual learning is in sound localization and segmentation \cite{liu2019automatic,liu2018automatic}. This task aims to predict the spatial location of individual sound sources in a video. Audio-visual localization and segmentation can be challenging due to the complex nature of the problem, as audio is not naturally aligned with all objects that exist in the video. However, with the recent advances in deep learning, researchers have developed many effective methods for this task.

One approach~\cite{mo2023av} to audio-visual localization and segmentation is to learn cross-modal representations that can align audio and visual information. In Fig.~\ref{fig: audio_visual_SA}, AV-SAM leverages pixel-wise audio-visual fusion across audio and visual features from the pre-trained audio encoder and image encoder to aggregate cross-modal representations. Then, the aggregated cross-modal features are fed into the prompt encoder and mask decoder to generate the final audio-visual segmentation masks.

Another approach to audio-visual localization and segmentation is to use contrastive learning to learn cross-modal correspondences. For example, EZ-VSL~\cite{mo2022EZVSL} and SLAVC~\cite{mo2022SLAVC} both use contrastive learning to learn the correspondence between audio and visual features. EZ-VSL uses a two-stream network to learn audio-visual alignment, while SLAVC uses a self-supervised contrastive learning approach to learn the correspondence between audio and visual features.

In addition to sound localization and segmentation, there are many other applications of audio-visual learning, such as audio-visual spatialization~\cite{Morgado2018selfsupervised, Morgado2020learning}, audio-event localization~\cite{tian2018ave,liu2017inner}, and audio-visual parsing~\cite{tian2020avvp,mo2022multimodal} have been addressed in previous studies. 

Overall, audio-visual learning has become an increasingly important field in deep learning and has many applications in various domains. With the recent advances in deep learning, we can expect to see more innovative methods for joint audio-visual learning in the future.

\subsubsection{Multimodal Visualization and Open-Vocabulary Interactive Segmentation}
\begin{figure}[htbp]
\centering
 \includegraphics[width=1\linewidth]{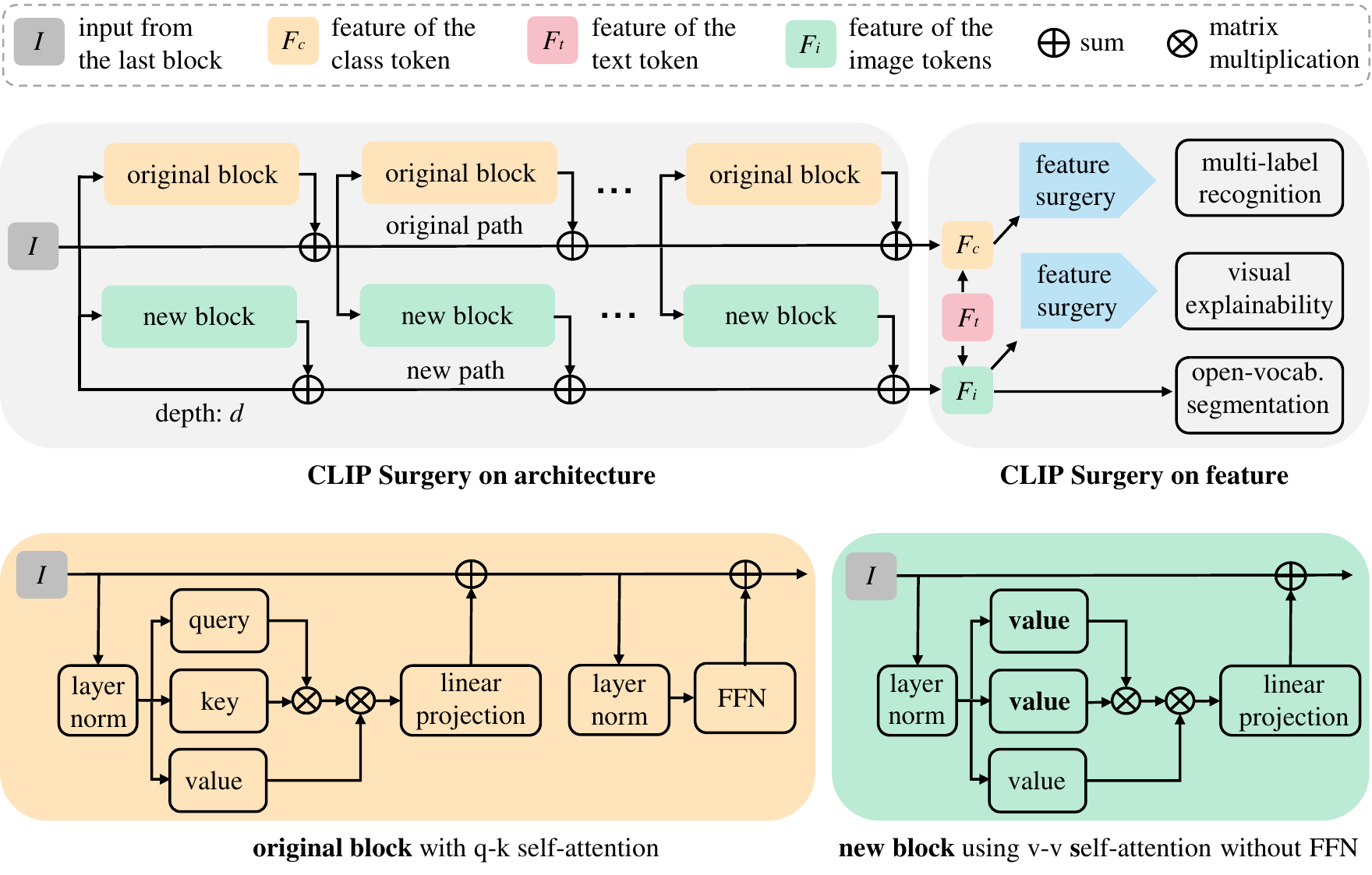}
\caption{Overview of the CLIP Surgery~\cite{clipsurgery2023}, which introduces dual paths for open-vocabulary interactive segmentation. The original path consists of the original block with q-k self-attention, while the new path incorporates the proposed v-v self-attention without a feed-forward network (FFN), starting from a certain depth, denoted as $d$. It is important to note that the parameters of the new block are copied from the original block to ensure consistency. Additionally, feature surgery is employed to combine the features of image tokens $\boldsymbol{F}_i$ with the text features $\boldsymbol{F}_t$ for the explainability task. Furthermore, feature surgery is applied to the feature of the class token $\boldsymbol{F}_c$ from the original path for multi-label recognition. The figure is borrowed from the original paper~\cite{clipsurgery2023}.}
\label{fig:archi_surgery}
\end{figure}

Recent work has shown that CLIP\cite{radford2021learning} can achieve impressive performance on a variety of vision tasks with minimal or no task-specific training. However, its internal mechanisms are not well understood. In this section, we discuss a recent work\cite{clipsurgery2023} that applies CLIP to the task of open-vocabulary interactive segmentation, which is closely related to the concept of multimodal visualization.

Interactive segmentation is a computer vision task that involves segmenting a target object from an image with user guidance in the form of points, scribbles, or boxes during the inference phase. The SAM \cite{sam} is a recent work that enables interactive segmentation via text prompts in an open-vocabulary manner. SAM requires manual points to guide the segmentation process.

The proposed method~\cite{wang2023caption} illustrated in Fig.~\ref{fig:archi_surgery} aims to replace the need for manual points entirely by using CLIP Surgery with text-only inputs. This approach provides pixel-level results from text input, which can be readily converted to point prompts for the SAM model. Specifically, the authors select foreground points ranked ahead in the similarity map and use the same number of points ranked last as background points. The authors show that their method outperforms other explainability methods in terms of both points accuracy and mIoU with SAM on four datasets.

The proposed method offers several advantages over other prompt formats in SAM. First, the method requires text input only, without the annotation cost of manual points suggested in the paper of SAM. Second, point prompts are superior to mask prompts because SAM's mask prompt is designed for its own output logits, and the generated points are more suitable than masks from another model. Finally, text-to-points are more readily achievable than the solution of text-to-boxes, which requires fine-tuning or additional supervision.

The proposed method also has implications for the explainability of CLIP in multimodal settings. Multimodal visualization is a promising direction for exploring CLIP's internal mechanisms. By visualizing the image-text pairs during training, the authors were able to observe interesting phenomena related to CLIP's learning process. However, the proposed method does not fully explain how CLIP is able to generate pixel-level results from text input. This suggests that further research is needed to better understand the mechanisms behind CLIP's impressive performance on open-vocabulary tasks.

In summary, the proposed method offers a promising solution for open-vocabulary interactive segmentation and has implications for the explainability of CLIP in multimodal settings. Future work could explore the potential of combining the proposed method with other SOTA models to achieve even better results.

\subsection{More Directions}
\subsubsection{Weakly-Supervised Semantic Segmentation} 
Recently, there are some works exploring the SAM in Weakly-Supervised Semantic Segmentation (WSSS). The advantage of using the SAM in WSSS is that it can produce satisfactory results without requiring model fine-tuning, which is a cumbersome requirement in current WSSS techniques such as \cite{song2019box,lee2021bbam}, which involve classification re-training and pseudo-label generation. By using SAM as a foundation model in WSSS, the process can be made more straightforward and less complex.

For example, 
\cite{sun2023alternative} aimed to investigate the suitability of the SAM for WSSS and adapt it to generate pseudo-labels using only image-level class labels. The study conducted an initial analysis of SAM as a foundation model in WSSS and discovered that it can achieve comparable performance without requiring fine-tuning. The study also revealed that SAM could generate high-quality segmentation masks and even surpass human annotations in some cases.

In \cite{sun2023alternative}, the performance of SAM was evaluated on the PASCAL VOC and MS-COCO datasets, where it demonstrated significant improvements over the latest SOTA methods. However, SAM encountered difficulties in certain situations due to the issue of semantic obscurity. While SAM performed well in most unambiguous settings, addressing semantic obscurity may require investigating the use of hierarchical-structured semantic classes and better prompts. Additionally, the study suggests exploring SAM's ability to segment "stuff" classes like "sky," "sea," and "road" to enhance overall scene understanding.
In \cite{jiang2023segment}, the authors presented a WSSS method that utilizes SAM as a pseudo-label generator. They employed diverse weak labels, such as image-level labels, points, scribbles, and bounding boxes, as prompts for SAM to produce precise class masks. These masks were then used to generate pseudo labels for training segmentation networks. The method's effectiveness was evaluated on the PASCAL VOC 2012 dataset, and the results indicate that SAM can serve as a reliable pseudo-label generator, with scribbles as prompts achieving an 89.7\% mIoU score on the training set. The final segmentation model obtained a 76.6\% mIoU score on the test set. This suggests that the method can be valuable in training segmentation networks with weak supervision, especially in scenarios where pixel-level annotations are unavailable.

However, in \cite{jiang2023segment}, the authors also pointed out some of SAM's limitations in handling WSSS. For instance, when using point prompts located by CAMs \cite{yang2019towards,liu2020semi}, SAM may generate incorrect object masks due to the CAMs' coarse locations. Additionally, in the case of using bounding box prompts, SAM may struggle to generate precise object masks when multiple objects are placed on the same table. These limitations imply that SAM may not always be capable of producing accurate object masks when dealing with WSSS.

In addition, due to the fact that current methods that rely on CAM to generate pseudo labels suffer from limitations such as partial and false activation. To overcome this, a new approach in \cite{chen2023segment} was introduced in this study that utilizes the SAM to generate high-quality pseudo labels by selecting relevant masks and labeling them based on initial CAM seed masks or post-processed pseudo labels. The results show that this approach significantly improves the precision of segmentation while also reducing computational costs compared to existing post-processing modules. The approach is highly versatile and compatible with existing WSSS models without modification to base networks or pipelines, and it has been shown to improve the mIoU of pseudo labels from five SOTA WSSS methods by an average of 6.2\% on the train set of the PASCAL VOC 2012 dataset.

\subsubsection{Adversarial Robustness}
As the concern regarding the susceptibility of computer vision systems to adversarial attacks grows \cite{wu2023adversarial}, there is a need to understand and research adversarial attacks. Adversarial attacks involve adding minor, undetectable perturbations to an input image to deceive the computer vision system into misclassifying the image \cite{zheng2022data,zheng2022pre}. This poses a significant concern because it may result in security breaches and safety risks in applications like autonomous vehicles \cite{cui2019review} and facial recognition systems \cite{anjos2011counter}.

Recently, \cite{zhang2023attack} assessed the adversarial robustness of the SAM model in the context of prompt-based segmentation in computer vision. The authors developed a framework called Attack-SAM to evaluate SAM's vulnerability to adversarial attacks in the prompt-based mask prediction task. This is the first comprehensive investigation of SAM's susceptibility to adversarial attacks, and the findings indicate that while SAM is vulnerable to white-box attacks, it remains somewhat robust in the black-box setting. Understanding the robustness of SAM in the face of adversarial attacks is crucial for developing more secure computer vision systems. The study provides the following insights: 
\begin{itemize}
    \item Firstly, the research reveals that SAM is vulnerable to white-box attacks, meaning that an attacker with full knowledge of the model's architecture and parameters can generate adversarial examples that deceive the model with ease. However, SAM is relatively robust in the black-box setting, where the attacker's knowledge of the model is limited. 
    \item Secondly, the study shows that small objects tend to be more resistant to adversarial attacks in the black-box setting, possibly due to the limited perturbation of the small object region. 
    \item Thirdly, the paper offers insights into the transferability of adversarial attacks among prompts in the segment everything mode of SAM. Additionally, the study provides a list of recent research papers and surveys on various topics related to generative AI, which can serve as a useful resource for researchers in the field.
\end{itemize}

Besides, \cite{guan2023badsam} presented BadSAM, which applied backdoor attacks to the SAM model for image segmentation. This paper demonstrated the potential security risks associated with such attacks and emphasized the need for further investigation into defense strategies in this area. The experiments conducted in the study demonstrate that the SAM model can be exploited by attackers, presenting a significant risk to end-users. The paper also outlines a process for initiating backdoor attacks on SAM and provides insights into the attacker's knowledge and methodology. In \cite{wang2023empirical}, the researchers carried out a thorough investigation into the robustness of SAM in various real-world scenarios. They conducted experiments that involved a wide range of image perturbations and domains, which revealed that SAM's performance typically deteriorates when dealing with perturbed images. However, the degree of vulnerability varies across different types of perturbations. By tailoring the prompting techniques and utilizing domain-specific knowledge that takes into account the unique characteristics of each dataset, it is possible to enhance the model's resilience to these perturbations and tackle challenges that are specific to each dataset. This work further highlights certain types of perturbations that have a significant impact on the model's ability to withstand challenges, identifying areas where improvements can be made.

\subsubsection{One Shot}
It was known that training the SAM model needs large datasets, which may be unreliable in some practical scenarios, such as medical imaging segmentation \cite{zhang2022hico}, or acoustic-to-articulatory analysis \cite{wang2023two}. As a result, there is a growing interest in exploring segmentation techniques that require minimal data and do not rely heavily on training. One of the benefits of utilizing SAM in one-shot learning is that it allows for personalized segmentation models to be efficiently created using only one-shot data, which consists of a user-provided image and a rough mask indicating the target object. To accomplish this, SAM's image encoder and the given mask are employed to encode the reference image's target object embedding. The feature similarity between the object and all the pixels on the new test image is then computed, and a positive-negative pair of points is selected, encoded as prompt tokens, and used as a location prior to SAM.

Recently, a novel training-free technique known as PerSAM was presented in \cite{zhang2023personalize} for creating personalized segmentation models for specific visual concepts using only one-shot data. PerSAM accomplishes this by localizing the target object in a reference image and utilizing target-guided attention, target-semantic prompting, and cascaded post-refinement to segment it in other images or videos. The study also introduces a variant of the approach, PerSAM-F, which fine-tunes only two parameters to enhance performance. The technique is assessed on a novel segmentation dataset, PerSeg, and is found to perform competitively in video object segmentation. The study also demonstrates that PerSAM can assist in improving personalized text-to-image generation.

\subsubsection{Explainable AI}
Explainable AI (XAI) plays a vital role in enhancing human comprehension of DNNs and has gained significant attention from researchers \cite{holzinger2022explainable}. It is worth noting that pixel-based XAI methods currently in use explain DNN decisions by identifying key pixels, while concept-based XAI methods aim to provide explanations using concepts. However, the interpretation of pixels can be challenging, and the imprecision of XAI methods can impact their reliability. Additionally, prior concept-based approaches either require human annotation or are limited to predefined concept sets.

\begin{figure}[htbp]
\centering
 \includegraphics[width=1\linewidth]{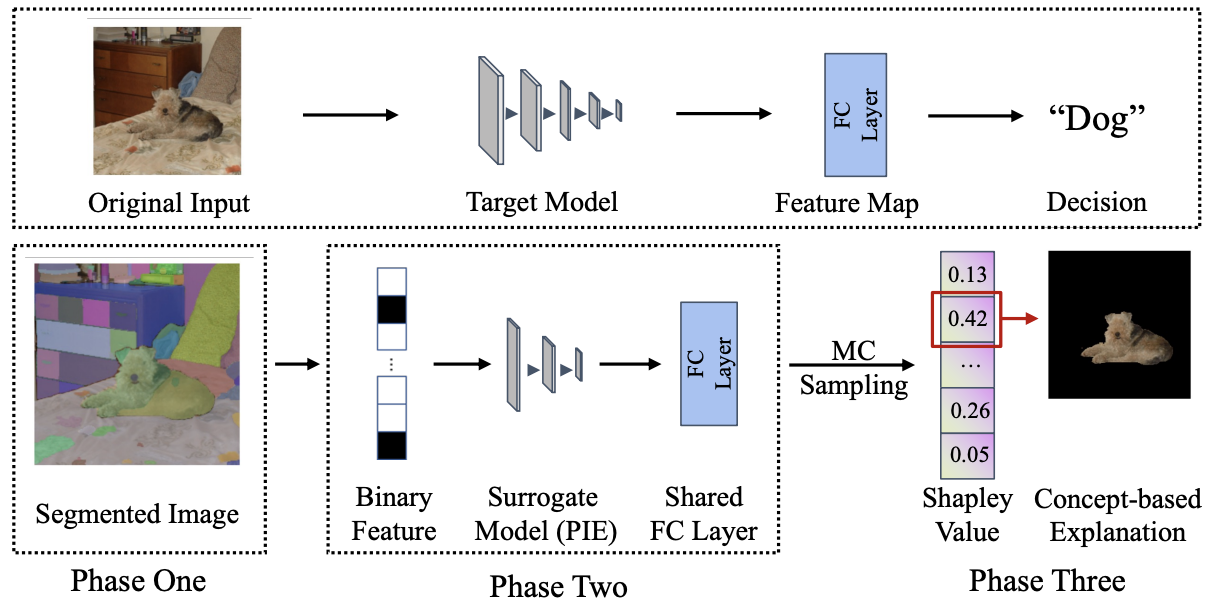}
\caption{Overview of the EAC approach, which employs a three-phase pipeline to provide explanations for a DNN’s prediction concerning an input image. More precisely, In the initial phase, it utilizes the widely accepted instance segmentation model, SAM, to partition the input image into a collection of visual concepts. Moving on to the second phase, a per-input equivalent (PIE) surrogate model is trained to approximate the behavior of the target DNN. Finally, in the third phase, the surrogate model is employed to efficiently explain the model’s prediction using the concepts obtained in the first phase. The figure is borrowed from the original paper~\cite{sun2023explain}.}
\label{fig:eac}
\end{figure}

A recent study \cite{sun2023explain} proposed an innovative approach that leverages the SAM to enhance concept-based XAI. The authors introduced Explain Any Concept (EAC), an effective and adaptable explanation method capable of elucidating DNN decisions using any concept. An overview of the EAC can be seen in Figure \ref{fig:eac}. Although SAM excels in instance segmentation, integrating it into XAI pipelines can be computationally demanding. To address this concern, the authors proposed a lightweight per-input equivalent (PIE) scheme, enabling efficient explanation using a surrogate model. Evaluation conducted on ImageNet and COCO datasets showcased the promising performance of EAC compared to commonly utilized XAI methods.

However, the paper raises a significant concern regarding the potential negative social impacts associated with using EAC in specific domains. It suggests that EAC could be misapplied to explain DNN predictions for medical images using unrelated concepts such as cats, trains, or subtle errors. Such misleading explanations could have severe consequences and misguide medical professionals in their decision-making process. Consequently, deploying EAC without adequate safety checks on its outputs in sensitive domains may pose significant risks. Future research efforts could focus on minimizing these negative societal impacts while exploring further applications of EAC in diverse visual domains and tasks.

\section{Conclusion}
\label{sec:conclusion}
This survey is the first to comprehensively review the recent progress on the foundation model of SAM for computer vision and beyond. Firstly, we summarize the development history of foundation models, covering large language models, large visual models, and large multimodal models, as well as the essential terminology about SAM. With a special focus on the applications of SAM to various tasks and data types, we summarize and compare the concurrent works of SAM and its follow-up works. Then, the great potential of SAM in a wide range of image processing applications is discussed, including software scenes, real-world scenes, and complex scenes. We also analyze and summarize the advantages and limitations of SAM across various applications. These observations can provide some insights to guide future research to develop stronger foundation models and further improve the robustness and generalization capabilities of SAM. Finally, we summarize massive other amazing applications of SAM in vision and beyond. {\color{black}The appendix provides a preliminary summary of open-source projects on SAM in table format.}


%

\ifCLASSOPTIONcaptionsoff
  \newpage
\fi



%

\bibliographystyle{IEEEtran}
\bibliography{ref}



%







\appendices
\renewcommand\thesection{\Alph{section}}
\setcounter{section}{0} 
\section{A Preliminary Summary of Open Source Projects on SAM}

\clearpage

{\scriptsize
\onecolumn
\begin{center}
    \begin{longtable}{|m{0.5cm}|m{1.5cm}|m{1.5cm}|m{3.5cm}|m{3.5cm}|m{1.5cm}|m{3cm}|} 
	\caption{Summary of Open Source Projects on SAM.} \label{tab:open_source_projects}  \\
         
	\hline 
        \multicolumn{1}{|c}{\textbf{No.}} & 
        \multicolumn{1}{|c}{\textbf{Project}} &
        \multicolumn{1}{|c}{\textbf{Title}} &
        \multicolumn{1}{|c}{\textbf{Project page}} &
        \multicolumn{1}{|c}{\textbf{Code base}} &
        \multicolumn{1}{|c}{\textbf{Affiliation}} & 
        \multicolumn{1}{|c|}{\textbf{Description}} \\ 
        \hline 
        \hline
	\endfirsthead 

	\multicolumn{3}{c}%
	{{\bfseries \tablename\ \thetable{} -- continued from previous page}} \\
	\hline 
        \multicolumn{1}{|c}{\textbf{No.}} & 
        \multicolumn{1}{|c}{\textbf{Project}} &
        \multicolumn{1}{|c}{\textbf{Title}} &
        \multicolumn{1}{|c}{\textbf{Project page}} &
        \multicolumn{1}{|c}{\textbf{Code base}} &
        \multicolumn{1}{|c}{\textbf{Affiliation}} & 
        \multicolumn{1}{|c|}{\textbf{Description}} \\ 
        \hline 
        \hline
	\endhead  
			
	\hline \multicolumn{3}{l}{{Continued on next page}} \\ 
	\endfoot  
			
	\hline
	\endlastfoot  

    001 & SAM & Segment Anything & \url{https://segment-anything.com/}  & \url{https://github.com/facebookresearch/segment-anything}   &  Meta   &   A foundation model for general segmentation.  \\
    
    \hline
    002 & SAM-Track & Segment and Track Anything & \url{https://colab.research.google.com/drive/1R10N70AJaslzADFqb-a5OihYkllWEVxB?usp=sharing} & \url{https://github.com/z-x-yang/Segment-and-Track-Anything}  &  Zhejiang University   &   A  project dedicated to tracking and segmenting any objects in videos, either automatically or interactively. \\
    
    \hline
    003 & Grounded-SAM & Grounded-Segment-Anything & \url{https://github.com/camenduru/grounded-segment-anything-colab}  & \url{https://github.com/IDEA-Research/Grounded-Segment-Anything}   & IDEA-Research  &   A project by combining Grounding DINO and SAM which aims to detect and segment Anything with text inputs. \\
    
    \hline 
    004 & MMDet-SAM & - & -  & \url{https://github.com/open-mmlab/playground/tree/main/mmdet_sam}   &  OpenMMLab   &   A new way of instance segmentation by combining SAM with Closed-Set Object Detection, Open-Vocabulary Object Detection, Grounding Object Detection. \\
    
    \hline 
    005 & MMRotate-SAM & Zero-shot Oriented Object Detection with SAM & -  & \url{https://github.com/open-mmlab/playground/tree/main/mmrotate_sam}   &  OpenMMLab   &   A project joins SAM and weakly supervised horizontal box detection to achieve rotated box detection. \\
    
    \hline 
    006 & MMOCR-SAM & - & -  & \url{https://github.com/open-mmlab/playground/tree/main/mmocr_sam}   &  OpenMMLab   &   A solution of Text Detection/Recognition and SAM that segments every text character, with striking text removal and text inpainting demos driven by diffusion models and Gradio. \\
    
    \hline 
    007 & MMEditing-SAM & - & -  & \url{https://github.com/open-mmlab/playground/tree/main/mmagic_sam}   &  OpenMMLab   &   A project join SAM and image generation to create awesome images and edit any part of them. \\
    
    \hline 
    008 & Label-Studio-SAM & OpenMMLab PlayGround: Semi-Automated Annotation with Label-Studio and SAM & -  & \url{https://github.com/open-mmlab/playground/tree/main/label_anything}   &  OpenMMLab    &   A project combining Label-Studio and SAM to achieve semi-automated annotation. \\
    
    \hline 
    009 & PaddleSeg & Segment Anything with PaddleSeg & -  & \url{https://github.com/PaddlePaddle/PaddleSeg/tree/release/2.8/contrib/SegmentAnything}  &  PaddlePaddle  &   A pretrained model parameters of PaddlePaddle format.\\
    
    \hline 
    010 & SegGPT & Segmenting Everything In Context & \url{https://huggingface.co/spaces/BAAI/SegGPT}  & \url{https://github.com/baaivision/Painter}   &  BAAI-Vision   &   SAM In Context based on Painter. \\
    
    \hline 
    011 & SEEM & Segment Everything Everywhere All at Once & \url{https://huggingface.co/spaces/xdecoder/SEEM}  & \url{https://github.com/UX-Decoder/Segment-Everything-Everywhere-All-At-Once}   &  Microsoft   &   A project can Segment Everything Everywhere with Multimodal prompts all at once. \\
    
    \hline 
    012 & CLIP Surgery & CLIP Surgery for Better Explainability with Enhancement in Open Vocabulary Tasks  & \url{https://github.com/xmed-lab/CLIP_Surgery/blob/master/demo.ipynb}  & \url{https://github.com/xmed-lab/CLIP_Surgery}   &  HKUST   &   A work about SAM based on CLIP's explainability to achieve text to mask without manual points. \\
    
    \hline 
    013 & SAMCOD & Can SAM Segment Anything? When SAM Meets Camouflaged Object Detection &  -  & \url{https://github.com/luckybird1994/SAMCOD}   &  -   &   SAM + Camouflaged object detection (COD) task. \\
    
    \hline 
    014 & Inpaint Anything & Segment Anything Meets Image Inpainting & \url{https://huggingface.co/spaces/InpaintAI/Inpaint-Anything} & \url{https://github.com/geekyutao/Inpaint-Anything}   &  USTC and EIT   &   SAM combines Inpainting, which is able to remove the object smoothly. \\
    
    \hline 
    015 & PerSAM & Personalize Segment Anything Model with One Shot & \url{https://huggingface.co/papers/2305.03048}  & \url{https://github.com/ZrrSkywalker/Personalize-SAM}   &   -  & SAM with specific concepts.  \\
    
    \hline 
    016 & MedSAM & Segment Anything in Medical Images
     & -  & \url{https://github.com/bowang-lab/MedSAM}   &  -   &   A step-by-step tutorial with a small dataset to help you quickly utilize SAM. \\
     
    \hline 
    017 & 
    Segment-Any-Anomaly & GroundedSAM Anomaly Detection & \url{https://colab.research.google.com/drive/1Rwio_KfziuLp79Qh_ugum64Hjnq4ZwsE?usp=sharing} & \url{https://github.com/caoyunkang/Segment-Any-Anomaly}   &  HUST  &   Grounding DINO + SAM to segment any anomaly. \\
    
    \hline 
    018 &  SSA & Semantic Segment Anything & -  & \url{https://github.com/fudan-zvg/Semantic-Segment-Anything}   &  Fudan University   &   A dense category annotation engine. \\
    
    \hline 
    019 & Magic Copy & - & -  & \url{https://github.com/kevmo314/magic-copy}   &  -  &   Magic Copy is a Chrome extension that uses SAM to extract a foreground object from an image and copy it to the clipboard. \\
    
    \hline 
    020 & Segment Anything with Clip & Segment Anything with Clip & \url{https://huggingface.co/spaces/curt-park/segment-anything-with-clip}  & \url{https://github.com/Curt-Park/segment-anything-with-clip}   &  -   &   SAM combined with CLIP. \\
    
    \hline 
    021 & MetaSeg & Segment Anything Video
     & \url{https://huggingface.co/spaces/ArtGAN/Segment-Anything-Video}  & \url{https://github.com/kadirnar/segment-anything-video}  &  -   &  Packaged version of the SAM. \\
     
    \hline 
    022 & SAM in Napari & Segment Anything Model (SAM) in Napari & \url{https://www.napari-hub.org/plugins/napari-sam}  & \url{https://github.com/MIC-DKFZ/napari-sam}   &  Applied Computer Vision Lab and German Cancer Research Center  &   Extended SAM's click-based foreground separation to full click-based semantic segmentation and instance segmentation. \\
    
    \hline 
    023 & SAM Medical Imaging & SAM Medical Imaging & -  & \url{https://github.com/amine0110/SAM-Medical-Imaging}   &  -   &   SAM for Medical Imaging. \\
    
    \hline 
    024 & 3D-Box & 3D-Box via Segment Anything & -  & \url{https://github.com/dvlab-research/3D-Box-Segment-Anything}   &  -   &   SAM is extended to 3D perception by combining it with VoxelNeXt. \\

    \hline 
    025 & Anything-3D & - & -  & \url{https://github.com/Anything-of-anything/Anything-3D}   &  -   &   Anything 3DNovel View, Anything-NeRF, Any 3DFace. \\
    
    \hline 
    026 & L2SET & Learning to Segment EveryThing &  -  & \url{https://github.com/ronghanghu/seg_every_thing}  & UC Berkeley, FAIR   &  A new partially supervised training paradigm for instance segmentation. \\
    
    \hline 
    027 & Edit Anything & Edit Anything by Segment-Anything & -  & \url{https://github.com/sail-sg/EditAnything}   &  -   &   Edit anything in images powered by SAM, ControlNet, StableDiffusion, \etc. \\
    
    \hline 
    028 & Image Edit Anything & IEA: Image Editing Anything & -  & \url{https://github.com/feizc/IEA}   &  -   & Using stable diffusion and SAM for image editing.   \\
    
    \hline 
    029 & SAM for Stable Diffusion Webui & Segment Anything for Stable Diffusion WebUI & -  & \url{https://github.com/continue-revolution/sd-webui-segment-anything}  &  -   &   This extension aim for connecting AUTOMATIC1111 Stable Diffusion WebUI and Mikubill ControlNet Extension with SAM and GroundingDINO to enhance Stable Diffusion/ControlNet inpainting. \\
    
    \hline 
    030 & Earth Observation Tools & Segment Anything EO tools & \url{https://colab.research.google.com/drive/1RC1V68tD1O-YissBq9nOvS2PHEjAsFkA?usp=share_link}  & \url{https://github.com/aliaksandr960/segment-anything-eo}   &  -   &   An earth observation tools for SAM. \\
    
    \hline 
    031 & Moving Object Detection & Towards Segmenting Anything That Moves & -  & \url{https://github.com/achalddave/segment-any-moving}   &  -   &   A project about SAM + Moving Object Detection. \\
    
    \hline 
    032 & OCR-SAM & Optical Character Recognition with Segment Anything & \url{https://www.zhihu.com/question/593914819/answer/2976012032}  & \url{https://github.com/yeungchenwa/OCR-SAM}  &  -   &  
    Combining MMOCR with SAM and Stable Diffusion. \\
    
    \hline 
    033 & SALT & Segment Anything Labelling Tool & -  & \url{https://github.com/anuragxel/salt#segment-anything-labelling-tool-salt}   &  -   &  A project uses the SAM Model and adds a barebones interface to label images and saves the masks in the COCO format. \\
    
    \hline 
    034 & Prompt Segment Anything & Prompt Segment Anything & -  & \url{https://github.com/RockeyCoss/Prompt-Segment-Anything}   &  -   &   An implementation of zero-shot instance segmentation using SAM. \\
    
    \hline 
    035 & SAM-RBox & - & -  & \url{https://github.com/Li-Qingyun/sam-mmrotate}   &  -   &  A project uses SAM for generating rotated bounding boxes with MMRotate, which is a comparison method of H2RBox-v2. \\
    
    \hline 
    036 & VISAM & MOTRv2: Bootstrapping End-to-End Multi-Object Tracking by Pretrained Object Detectors & -  & \url{https://github.com/BingfengYan/VISAM}  &  - &  Combining SAM with MOT, it creates the era of "MOTS". \\
    
    \hline 
    037 & SegEO & Segment Anything EO tools & -  & \url{https://github.com/aliaksandr960/segment-anything-eo}   &  -   &   The tools are developed to ease the processing of spatial data (GeoTIFF and TMS) with SAM using sliding window algorithm for big files. \\
    
    \hline 
    038 & Napari Segment Anything
     & Napari Segment Anything & \url{https://app.codecov.io/gh/jookuma/napari-segment-anything}  & \url{https://github.com/JoOkuma/napari-segment-anything}   &  -   &   SAM native Qt UI. \\
     
    \hline 
    039 & Segment-Anything-U-Specify & Segment-Anything-U-Specify & -  & \url{https://github.com/MaybeShewill-CV/segment-anything-u-specify}   &  -   & Using CLIP and SAM to segment any instance you specify with text prompt of any instance names. \\
    
    \hline 
    040 & SegDrawer & Simple static web-based mask drawer & \url{https://colab.research.google.com/drive/1PdWCpBgYwiQtvkdTBnW-y2T-s_Fc-2iI?usp=sharing} & \url{https://github.com/lujiazho/SegDrawer}   &  -   &   Simple static web-based mask drawer, supporting semantic segmentation with SAM. \\
    
    \hline 
    041 & Track Anything & Segment Anything Meets Videos & \url{https://huggingface.co/spaces/VIPLab/Track-Anything}  & \url{https://github.com/gaomingqi/Track-Anything}   &  SUSTech   &  Track-Anything is a flexible and interactive tool for video object tracking and segmentation. \\
    
    \hline 
    042 & Count Anything & - & -  & \url{https://github.com/ylqi/Count-Anything}  &  -   &   A method uses SAM and CLIP to ground and count any object that matches a custom text prompt, without requiring any point or box annotation. \\
    
    \hline 
    043 & RAM & Relate Anything Model  & \url{https://huggingface.co/spaces/mmlab-ntu/relate-anything-model}  & \url{https://github.com/Luodian/RelateAnything}   &  MMLab, NTU and VisCom Lab, KCL/TongJi   &   Relate Anything Model is capable of taking an image as input and utilizing SAM to identify the corresponding mask within the image. \\
    
    \hline 
    044 & Segment Any RGBD & Segment Any RGBD & \url{https://github.com/Jun-CEN/SegmentAnyRGBD}  &  \url{https://huggingface.co/spaces/jcenaa/Segment-Any-RGBD}  &  -   &  Segment AnyRGBD is a toolbox to segment rendered depth images based on SAM. \\
    
    \hline 
    045 & Show Anything & Show Anything & \url{https://huggingface.co/spaces/weijiawu/ImageEditAnything} & \url{https://github.com/showlab/ShowAnything}   &  Showlab, NUS   &   Some Applications that are compatible with both SAM and Generation. \\
    
    \hline 
    046 & Transfer Any Style
     & Any-to-Any Style Transfer: Making Picasso and Da Vinci Collaborate & -  & \url{https://github.com/Huage001/Transfer-Any-Style}   &  LV-lab, NUS   &  An interactive demo based on Segment-Anything for style transfer which enables different content regions apply different styles. \\
     
    \hline 
    047 & Caption Anything & - & \url{https://colab.research.google.com/github/ttengwang/Caption-Anything/blob/main/notebooks/tutorial.ipynb} &  \url{https://github.com/ttengwang/Caption-Anything}  & VIP lab, SUSTech  &   Caption-Anything is a versatile image processing tool that combines the capabilities of SAM, Visual Captioning, and ChatGPT.  \\
    
    \hline 
    048 & Image2Paragraph & Transform Image Into Unique Paragraph & \url{https://zhaohengyuan1.github.io/image2paragraph.github.io/} & \url{https://github.com/showlab/Image2Paragraph} & - & Transform Image into Unique Paragraph with ChatGPT, BLIP2, OFA, GRIT, Segment Anything, ControlNet. \\
    
    \hline 
    049   & LIME SAM & Local Interpretable Model-agnostic Explanations Segment Anything & \url{https://colab.research.google.com/drive/1bj6B-O47NHpqsWovOrVZcpWNhIfO56sj?usp=sharing} & \url{https://github.com/jaydeep-work/LIME-SAM} & - &LIME-SAM aims to create an Explainable Artificial Intelligence (XAI) framework for image classification using LIME (Local Interpretable Model-agnostic Explanations) as the base algorithm, with the super-pixel method replaced by SAM. \\
    
    \hline 
    050 & Paint Anything & - & - & \url{https://github.com/Huage001/Paint-Anything} & - & An interactive demo based on SAM for stroke-based painting which enables human-like painting. \\
    
    \hline 
    051 & SAMed & Customized Segment Anything Model for Medical Image Segmentation & \url{https://colab.research.google.com/drive/1KCS5ulpZasYl9DgJJn59WsGEB8vwSI_m?usp=sharing} & \url{https://github.com/hitachinsk/SAMed} & USTC & SAMed is built upon the large-scale image segmentation model, SAM, to explore the new research paradigm of customizing large-scale models for medical image segmentation. \\
    
    \hline 
    052 & Personalize SAM
     & Personalize Segment Anything with 1 Shot in 10 Seconds & \url{https://huggingface.co/spaces/justin-zk/Personalize-SAM} & \url{https://github.com/ZrrSkywalker/Personalize-SAM} & MMLab, CUHK & A training-free Personalization approach for SAM, termed as PerSAM. Given only a single image with a reference mask, PerSAM can segment specific visual concepts. \\
     
    \hline 
    053  &  Open-vocabulary-Segment-Anything  & Open-vocabulary-Segment-Anything  &  -  &  \url{https://github.com/ngthanhtin/owlvit_segment_anything}  &  - &  Combining OwlViT with Segment Anything - Open-vocabulary Detection and Segmentation (Text-conditioned, and Image-conditioned).\\

     \hline 
    054  &  Labal-Anything-Pipeline & Label-Anything-Pipeline &  -  & \url{https://github.com/Yuqifan1117/Labal-Anything-Pipeline} & ZJU  & Annotation anything in visual tasks just all in one-pipeline with GPT-4. and SAM. \\

    \hline 
    055  &  Grounded-Segment-Any-Parts  &  Grounded Segment Anything: From Objects to Parts  &  \url{https://cheems-seminar.github.io/}  &  \url{https://github.com/Cheems-Seminar/grounded-segment-any-parts}  &  HKU  &  Expand Segment Anything Model (SAM) to support text prompt input. The text prompt could be object-level(eg, dog) and part-level(eg, dog head).  \\

     \hline 
    056  &  AnyLabeling  & AnyLabeling  &  \url{https://www.youtube.com/watch?v=5qVJiYNX5Kk}  &  \url{https://github.com/vietanhdev/anylabeling}  &  -  &  Effortless AI-assisted data labeling with AI support from Segment Anything and YOLO. \\

     \hline 
    057  & SSA &  Semantic-Segment-Anything  &  \url{https://replicate.com/cjwbw/semantic-segment-anything}  &  \url{https://github.com/fudan-zvg/Semantic-Segment-Anything}  &  -  &  Automated dense category annotation engine that serves as the initial semantic labeling for the Segment Anything dataset (SA-1B).\\ 

     \hline 
    058  & RefSAM &  Label Data with Segment Anything in Roboflow  &  \url{https://blog.roboflow.com/label-data-segment-anything-model-sam/}  &  \url{https://github.com/helblazer811/RefSAM}  &  -  &  Referring Image Segmentation Benchmarking with Segment Anything Model (SAM). \\

     \hline 
    059  & Roboflow Annotate &  Launch: Label Data with Segment Anything in Roboflow  &  \url{https://blog.roboflow.com/label-data-segment-anything-model-sam/}  &  \url{https://app.roboflow.com/}  &  Roboflow  &  SAM-assisted labeling for training computer vision models. \\

     \hline 
    060  & ImageBind SAM &  -  &  -  & \url{https://github.com/IDEA-Research/Grounded-Segment-Anything/tree/main/playground/ImageBind_SAM}  &  IDEA-Research  &  This is an experimental demo aims to combine ImageBind and SAM to generate mask with different modalities. \\

    

    \end{longtable}
\end{center}
}

\end{document}